%% file: main.tex
\documentclass[runningheads]{llncs}

\usepackage{eccv}

\usepackage{eccvabbrv}

\usepackage{graphicx}
\usepackage{booktabs}
\usepackage{caption} 
\usepackage{float}
\usepackage[accsupp]{axessibility}  % Improves PDF readability for those with disabilities.

\usepackage{hyperref}

\usepackage{orcidlink}

\begin{document}

% ---------------------------------------------------------------
% TODO REVIEW: Replace with your title
\title{CalliMaster: Mastering Page-level Chinese Calligraphy via Layout-guided Spatial Planning} 

% TODO REVIEW: If the paper title is too long for the running head, you can set
% an abbreviated paper title here. If not, comment out.
\titlerunning{CalliMaster}

\author{\small Tianshuo Xu$^1$\and Tiantian Hong$^2$\and Zhifei Chen$^1$\and Fei Chao$^3$\and Ying-cong Chen$^{1,4,*}$ \\
% \url{https://github.com/Tianshuo-Xu/Motion-Forcing} 
}

% TODO FINAL: Replace with an abbreviated list of authors.
\authorrunning{T. Xu et al.}
% First names are abbreviated in the running head.
% If there are more than two authors, 'et al.' is used.

% TODO FINAL: Replace with your institution list.
\institute{The Hong Kong University of Science and Technology (Guangzhou)\and
Faculty of Engineering and IT, University of Technology Sydney\and
Xiamen University\and
The Hong Kong University of Science and Technology \\
\email{\{txu647\}@connect.hkust-gz.edu.cn}}

\maketitle

\begin{center}
    \vspace{-2em}
    \includegraphics[width=0.9\textwidth]{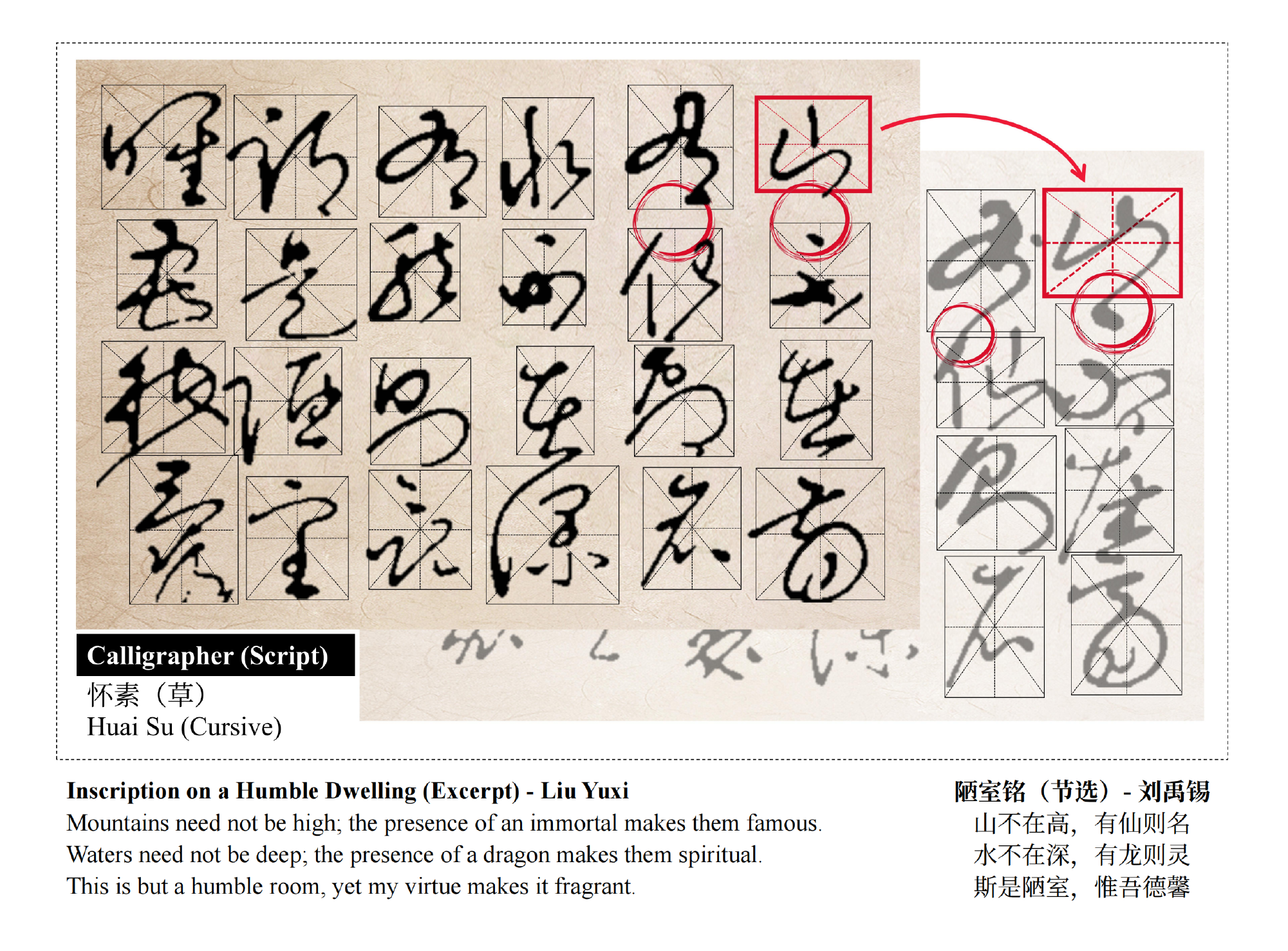}
    \vspace{-1em}
    \captionof{figure}{\textbf{Page-Level Generation and Editing.} \textbf{(Left)} CalliMaster synthesizes high-fidelity Chinese calligraphy that harmonizes local glyph precision with global spatial rhythm. \textbf{(Right)} Interactive semantic re-planning. By adjusting bounding boxes (\textcolor{red}{$\square$}), users can modify the layout while the model regenerates continuous inter-character strokes (\textcolor{red}{$\bigcirc$}) and optimizes the surrounding void space to preserve artistic momentum.}
    \label{fig:teaser}
\end{center}
\begin{abstract}
Page-level calligraphy synthesis requires balancing glyph precision with layout composition. Existing character models lack spatial context, while page-level methods often compromise brushwork detail. In this paper, we present \textbf{CalliMaster}, a unified framework for controllable generation and editing that resolves this conflict by decoupling spatial planning from content synthesis. Inspired by the human cognitive process of ``planning before writing'', we introduce a coarse-to-fine pipeline \textbf{(Text $\rightarrow$ Layout $\rightarrow$ Image)} to tackle the combinatorial complexity of page-scale synthesis. Operating within a single Multimodal Diffusion Transformer, a spatial planning stage first predicts character bounding boxes to establish the global spatial arrangement. This intermediate layout then serves as a geometric prompt for the content synthesis stage, where the same network utilizes flow-matching to render high-fidelity brushwork. 
Beyond achieving state-of-the-art generation quality, this disentanglement supports versatile downstream capabilities. By treating the layout as a modifiable constraint, CalliMaster enables controllable semantic re-planning: users can resize or reposition characters while the model automatically harmonizes the surrounding void space and brush momentum. Furthermore, we demonstrate the framework's extensibility to artifact restoration and forensic analysis, providing a comprehensive tool for digital cultural heritage.

\end{abstract}

\section{Introduction}
\label{sec:intro}

Chinese calligraphy is practiced by millions worldwide, revered for its profound historical depth and artistic sophistication~\cite{casti_art_exam_1_8m_2022, nihonshuji_students_300k_2025}. However, replicating this complexity is difficult because calligraphy is essentially a minimalist line art that captures the continuous kinetics of the writer. Every stroke serves as a skeletal record of movement, where the ``negative space'' (the white void) carries as much structural weight as the black ink. Generating a calligraphic page necessitates orchestrating two distinct motion modalities~\cite{chen2017calligraphy, qiu2022calligraphy}: the external trajectory of the brush, governed by the artist's anticipatory spatial planning, which establishes the global layout and spatial boundaries, and its internal modulation, where subtle tip manipulations dictate microscopic texture and ink variations. To date, deep learning has struggled to reconcile these conflicting objectives, often forcing a trade-off between scale and fidelity.
Early research focused on \textbf{isolated character synthesis}~\cite{Zhang2024DPFont, yang2024fontdiffuser, Xie2021DGFont}, achieving high glyph accuracy but treating the layout as a rigid grid, thereby severing the crucial inter-character ligatures and destroying the spatial rhythm. Recently, UniCalli~\cite{xu2025unicalli} pioneered the extension of precise generation to the column level, attempting to capture vertical flow. However, extending this capability from linear vertical columns to complex spatial page layouts remains elusive. 

This transition induces an \textbf{exponential expansion} in complexity. Mathematically, the spatial interdependence between characters transforms the synthesis task from a series of independent local predictions into a high-dimensional \textit{joint} optimization problem. Unlike isolated generation where the solution space is constrained locally, page-level synthesis confronts a combinatorial explosion: the model must optimize fine-grained stylistic texture for individual glyphs (local constraints) while strategically planning void space to satisfy the global equilibrium of the layout. This computational intractability stems directly from the artistic necessity of modeling dynamic momentum rather than static shapes. To capture the artist's overarching intention, the model must synthesize the ``external trajectory''---the invisible kinematic link connecting the terminus of one glyph to the inception of the next—which dictates the rhythmic flow of the entire page. Current general-purpose models are ill-equipped to manage this dual-objective optimization. While Large Vision Models (LVMs)~\cite{labs2025flux1kontextflowmatching} and Vision-Language Models (VLMs)~\cite{qwen2025vl} can generate coherent compositions, they lack domain-specific priors, frequently yielding hallucinated structures or generic, ``font-like'' strokes that fail to capture the nuances of specific artistic styles. Consequently, generating an artistically authentic, structurally accurate calligraphic page remains a significant open challenge.

The imperative for solving this challenge extends beyond academic preservation to the frontier of modern digital creativity. With the resurgence of ``Neo-traditional'' aesthetics in gaming and design, there is a critical demand for customizable calligraphic assets. In these dynamic workflows, static generation is insufficient. Designers require the flexibility to manipulate the layout, such as resizing a title character or reflowing a couplet, without breaking the artistic coherence. This necessitates a model capable of ``semantic re-planning''. Specifically, when characters or layouts are modified, the system must redesign the spatial gaps and brush momentum rather than merely inpainting pixels. Existing editing paradigms fail to meet this demand. Simple warping distorts the rigorous geometry of brushstrokes, while standard inpainting operates within fixed boundaries, failing to propagate local changes to the global layout. Without the capacity for holistic spatial readjustment, these methods inevitably sever the rhythmic connection between characters, fracturing the visual flow.
% Existing pixel-level editing algorithms (e.g., standard inpainting or warping) fail here because they treat the canvas as a static grid of pixels rather than a continuous sequence of brush movements, inevitably fracturing the visual flow.

To bridge this gap, we propose \textbf{CalliMaster}, a unified generative framework that reformulates page-level synthesis by mirroring the human cognitive process of ``planning before writing''. Positing that the limitations of existing methods stem from the entanglement of macro-scale composition with micro-scale rendering, we systematically decouple the generation process into two interdependent stages: \textbf{spatial planning} and \textbf{content synthesis}. 
Technically, we materialize this through a \textbf{single unified Multimodal Diffusion Transformer}, where Text, Layout, and Image are modeled as a continuous sequence of tokens governed by independent noise schedules. Within this framework, the generation proceeds in a coarse-to-fine manner. In the initial \textbf{spatial planning stage}, the model functions as a global planner, predicting a layout (represented as character bounding boxes) to establish the spatial coherence. By optimizing position and scale to harmonize density and void, this stage effectively digitizes the artist's anticipatory intention. The resulting layout then serves as a geometric prompt for the subsequent \textbf{content synthesis stage}, where the \textit{same} network utilizes flow-matching to render high-fidelity brushwork within the defined regions. By strictly enforcing this causal dependency (Text $\rightarrow$ Layout $\rightarrow$ Image), CalliMaster ensures that the stochastic generation of texture remains anchored to a coherent structural skeleton.

Beyond standard generation, this architectural disentanglement fundamentally transforms the paradigm of interaction and application. By leveraging the flexible inference capabilities inherent to this probabilistic framework, CalliMaster demonstrates robust performance across a spectrum of downstream tasks. In the realm of interactive editing, user manipulations of the layout function as explicit \textbf{geometric prompts}. These prompts compel the model to perform \textbf{semantic re-planning}: re-inferring optimal brush trajectories and inter-character ligatures conditioned on modified spatial constraints, thereby preserving artistic integrity during redesign. We extend this capability to artifact restoration, where the model utilizes layout-aware priors to reconstruct fractured strokes by inferring missing motion flows from fragmentary evidence. Additionally, we pioneer a forensic application for calligraphy identification, analyzing the consistency of diffusion denoising likelihood to quantify stylistic deviations. This demonstrates that CalliMaster is not merely a generation tool, but a comprehensive platform for the creation, restoration, and analysis of calligraphic art.

In summary, our main contributions are as follows:
\begin{itemize}
    \item \textbf{Unified Page-Level Generation Framework:} We propose CalliMaster, a novel pre-trained large Calligraphy Model that decouples spatial planning from content synthesis. By leveraging a Diffusion Forcing-inspired architecture with a coarse-to-fine pipeline (Text $\rightarrow$ Layout $\rightarrow$ Image), we achieve high-fidelity page-level generation that simultaneously preserves local glyph accuracy and global brush momentum.
    
    \item \textbf{Semantic Layout Editing via Geometric Prompts:} We introduce a box-guided interaction mechanism that transforms editing from rigid pixel manipulation into semantic re-planning. This allows users to modify layout structures (e.g., shifting or resizing characters) while the model regenerates the necessary inter-character ligatures, surrounding negative space, and spatial balance to maintain artistic integrity.
    
    \item \textbf{Extensible Applications for Cultural Heritage:} We demonstrate the model's versatility in downstream tasks beyond generation. Specifically, we apply CalliMaster to \textbf{artifact restoration} via layout-aware inpainting and propose a novel forensic metric based on \textbf{diffusion denoising likelihood}, offering a quantifiable approach for the identification and authentication of calligraphic works.
\end{itemize}

\section{Related Work}
\label{sec:related_work}

\subsection{From Isolated Characters to Page-Level Calligraphy}
Bridging structural fidelity and holistic aesthetics in calligraphy generation remains a significant challenge. Early GAN-based~\cite{wu2020calligan, Xie2021DGFont} and recent diffusion-based methods~\cite{Zhang2024DPFont, He2024DiffFontIJCV} excel at isolated character synthesis but inherently treat glyphs as independent icons. Consequently, they fail to model the spatial rhythm and the continuous \textbf{global brush momentum} that defines a complete work. To address this, systems like CalliPaint~\cite{Liao2023CalliPaint} and CalliffusionV2~\cite{Liao2024CalliffusionV2} attempt page-level synthesis via autoregressive paradigms. However, their unidirectional dependency prevents holistic spatial planning and renders interactive editing intractable. Most recently, UniCalli~\cite{xu2025unicalli} introduced unified column-level generation. Crucially, while it incorporates bounding boxes, it generates them simultaneously with the image pixels. This coupled approach treats layout as a parallel co-objective rather than a prerequisite plan, failing to decouple spatial planning from content synthesis. In contrast, CalliMaster employs a strict coarse-to-fine mechanism (Text $\to$ Layout $\to$ Image), where boxes serve as explicit geometric priors to guide high-fidelity brushwork generation.

\subsection{Sequence Modeling in Diffusion Frameworks}
Diffusion models~\cite{ho2020denoising, song2020denoising} dominate high-fidelity image synthesis, yet standard architectures often lack the precise spatial planning required for sequence generation. To address this, recent works have introduced mechanisms to enhance sequential reasoning. For instance, MMDiT~\cite{blattmann2023stable, labs2025flux1kontextflowmatching} employs bidirectional attention for multimodal sequences, while Diffusion Forcing~\cite{chen2024diffusion} bridges next-token prediction and diffusion by assigning independent, per-token noise levels. This paradigm allows for ``causal'' generation—planning future tokens based on past ones—without losing global context. Drawing inspiration from these advances, CalliMaster leverages independent noise schedules to enforce the causal dependency of our coarse-to-fine pipeline, ensuring the macro-level spatial plan is finalized before micro-level pixel synthesis.

\subsection{Layout-Aware Editing and Semantic Re-planning}
Editing calligraphic images requires understanding both textual content and artistic flow, a capability current general-purpose models lack. Scene text editors like AnyText~\cite{tuo2024anytext} and TextDiffuser-2~\cite{chen2023textdiffuser2unleashingpowerlanguage} treat text as rigid visual patterns rather than kinematic trajectories, failing to reconstruct the surrounding void space and inter-character ligatures when layouts are altered. Similarly, restoration models like DiffACR~\cite{li2024diffacr} focus on passive inpainting rather than active semantic re-planning. Our work bridges this gap by utilizing bounding boxes as geometric prompts. This allows users to manipulate the layout, compelling the model to automatically re-infer the continuous brush momentum and redesign the void space, thereby maintaining artistic coherence in the modified composition.

\section{Method}
\label{sec:method}

\begin{figure}[!t]
    \centering
    \includegraphics[width=\linewidth]{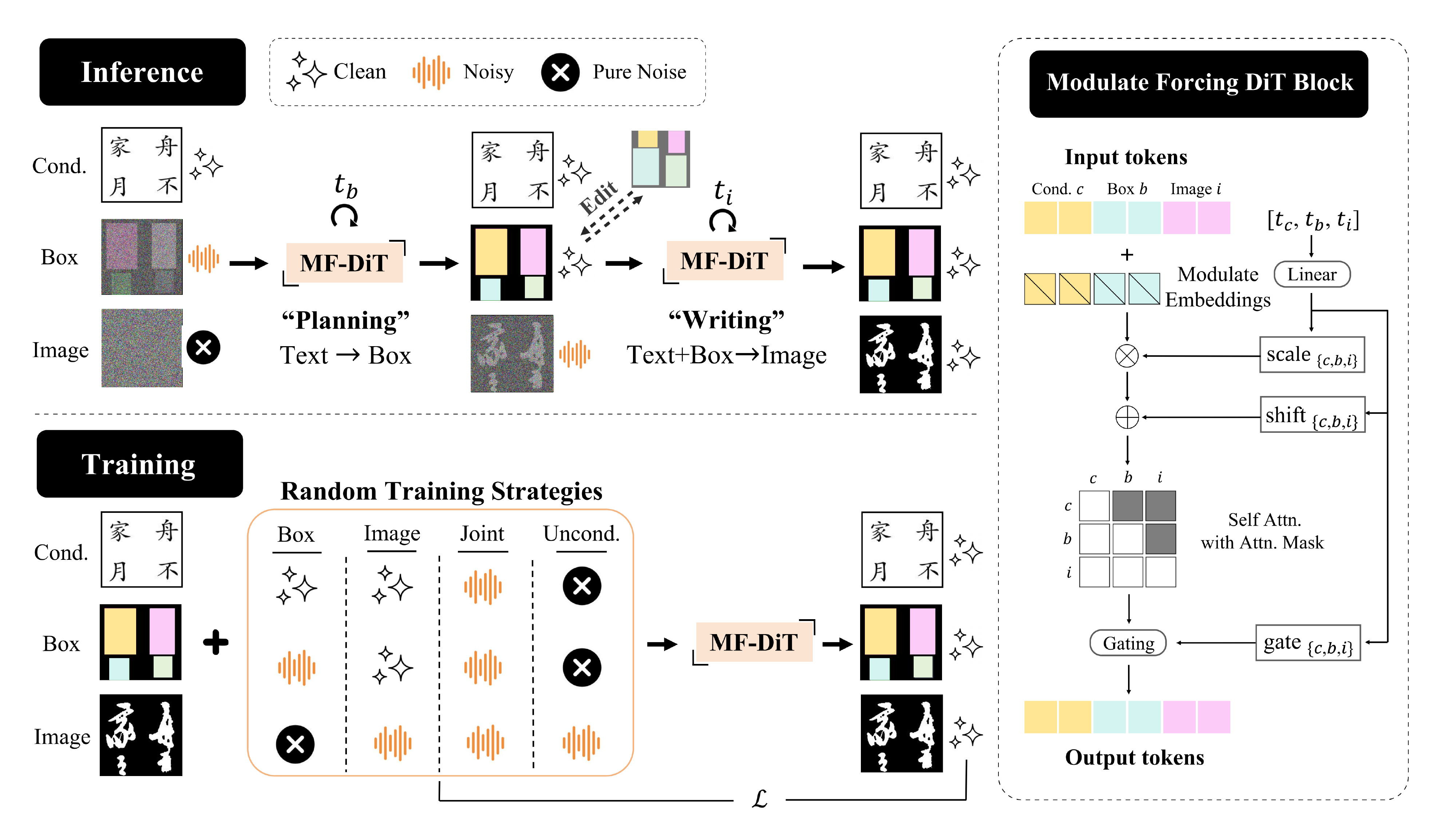}
    \caption{\textbf{Overview of CalliMaster.} The framework decouples calligraphy generation into two core stages: spatial planning and content writing. Inference executes these sequentially: planning the layout before writing the content. Training randomly samples between these two primary stages and two auxiliary states to jointly optimize the model. To enable this unified strategy, the core MF-DiT block employs modality-aware AdaLN driven by independent timesteps ($t_c$, $t_b$, $t_i$), structural attention masking, and modulate embeddings.}
    \label{fig:pipeline}
\end{figure}

We present CalliMaster (shown in \cref{fig:pipeline}), a unified framework for Chinese calligraphy generation and editing. The core idea is to decompose the complex page-level synthesis task into two tractable sub-problems: \textbf{spatial planning} and \textbf{content synthesis}. We achieve this by extending a rectified flow transformer with a Diffusion Forcing-inspired multi-timestep architecture. In this section, we describe the overall formulation (\cref{sec:formulation}), the model architecture (\cref{sec:architecture}), and the multi-objective training strategy (\cref{sec:training}). Finally, we detail the inference procedure and various downstream applications (\cref{sec:inference_apps}).
% including layout-aware editing and restoration (\cref{sec:inference_apps}).

\subsection{Problem Formulation}
\label{sec:formulation}

Given a text prompt $\mathbf{p}$ specifying the characters to render, a style description $\mathbf{s}$ (e.g., calligraphic script, author), and an auxiliary reference condition image $\mathbf{c}$ (providing foundational structural font priors) depicting standard-font glyphs in their intended positions, our goal is to generate a calligraphy image $\mathbf{x}$ that (i) faithfully renders the specified characters, (ii) captures the target artistic style, and (iii) preserves global spatial coherence, the rhythm of density and void that defines a calligraphic composition.

We adopt a \emph{rectified flow} formulation, a specific instantiation of \textbf{flow matching}~\cite{liu2022flow, esser2024scaling}. Let $\mathbf{x}_0$ denote the clean data and $\boldsymbol{\epsilon} \sim \mathcal{N}(\mathbf{0}, \mathbf{I})$ denote Gaussian noise. The forward interpolation at noise level $t \in [0, 1]$ is:
\begin{equation}
    \mathbf{x}_t = (1 - t)\,\mathbf{x}_0 + t\,\boldsymbol{\epsilon},
    \label{eq:interpolation}
\end{equation}
and the model is trained to predict the velocity field $\mathbf{v} = \boldsymbol{\epsilon} - \mathbf{x}_0$.

\paragraph{From Single to Multi-Timestep.}
Unlike standard flow matching where a single timestep $t$ governs the entire input, we introduce \emph{independent noise schedules} for three token groups (Image, Box, and Condition) inspired by Diffusion Forcing~\cite{chen2024diffusion}. We define a timestep triplet $(t_{\mathrm{img}}, t_{\mathrm{box}}, t_{\mathrm{cond}})$, where each component controls the noise level of its respective modality:
\begin{align}
    \mathbf{x}_t^{\mathrm{img}} &= (1 - t_{\mathrm{img}})\,\mathbf{x}_0 + t_{\mathrm{img}}\,\boldsymbol{\epsilon}_{\mathrm{img}}, \label{eq:noise_img}\\
    \mathbf{b}_t^{\mathrm{box}} &= (1 - t_{\mathrm{box}})\,\mathbf{b}_0 + t_{\mathrm{box}}\,\boldsymbol{\epsilon}_{\mathrm{box}}, \label{eq:noise_box}\\
    \mathbf{c}_t^{\mathrm{cond}} &= (1 - t_{\mathrm{cond}})\,\mathbf{c}_0 + t_{\mathrm{cond}}\,\boldsymbol{\epsilon}_{\mathrm{cond}}, \label{eq:noise_cond}
\end{align}
where $\mathbf{b}_0$ and $\mathbf{c}_0$ are the clean box layout and condition image latents, respectively.

\subsection{Model Architecture}
\label{sec:architecture}

We instantiate CalliMaster using a Multimodal Diffusion Transformer (MM-DiT) backbone~\cite{peebles2023scalable, labs2025flux1kontextflowmatching}. To accommodate the unique requirements of calligraphic generation, we extend the standard architecture to process heterogeneous modalities, including text, layouts, and visual conditions, within a unified sequence.

\paragraph{Token Sequence Construction.}
Given an input image $\mathbf{I} \in \mathbb{R}^{H \times W \times 3}$, we first encode it into a latent representation using a frozen VAE, followed by patchification with a patch size of $p \times p$. This yields a sequence of $N$ visual tokens. To unify multimodal inputs, we construct a composite sequence $\mathbf{Z}$ containing text, image, box layout, and condition tokens:
\begin{equation}
    \mathbf{Z} = [\,\underbrace{\mathbf{z}_{\mathrm{txt}}^{1:L} \vphantom{\mathbf{z}_{\mathrm{img}}^{1:N}}}_{\text{text}},\;\underbrace{\mathbf{z}_{\mathrm{img}}^{1:N}}_{\text{image}},\;\underbrace{\mathbf{z}_{\mathrm{box}}^{1:N} \vphantom{\mathbf{z}_{\mathrm{img}}^{1:N}}}_{\text{box layout}},\;\underbrace{\mathbf{z}_{\mathrm{cond}}^{1:N} \vphantom{\mathbf{z}_{\mathrm{img}}^{1:N}}}_{\text{condition}}\,],
    \label{eq:token_seq}
\end{equation}
where $L$ denotes the text sequence length. While text tokens are projected linearly from a text encoder embedding, the three visual streams ($\mathrm{img}, \mathrm{box}, \mathrm{cond}$) share a common input projection layer initialized with weights $W_{\mathrm{in}}$. To disambiguate these streams within the shared latent space, we introduce learnable \textbf{modality embeddings} $\mathbf{e}_m$ for $m \in \{\mathrm{img}, \mathrm{box}, \mathrm{cond}\}$, such that the input to the transformer is $\mathbf{z} = W_{\mathrm{in}}\,\mathbf{p}_{\mathrm{patch}} + \mathbf{e}_m$.

To further encode spatial correspondence and modality identity, we extend 3D Rotary Position Embeddings (RoPE)~\cite{su2024roformer} to jointly represent the modality index $m$ and 2D coordinates $(y,x)$, ensuring the model distinguishes overlapping visual streams while maintaining their spatial alignment.

\paragraph{Modality-Aware Adaptive Layer Normalization.}
A core contribution of our architecture is the decoupling of noise schedules across modalities. We implement this via \emph{modality-aware AdaLN}~\cite{peebles2023scalable}. Each modality $m$ is assigned an independent timestep $t_m$, which is mapped to a conditioning vector $\mathbf{h}_m$ via a shared timestep embedder:
\begin{equation}
    \mathbf{h}_m = \mathrm{MLP}_{\mathrm{time}}(\gamma(t_m)) + \mathrm{MLP}_{\mathrm{vec}}(\mathbf{y}_{\mathrm{txt}}),
    \label{eq:timestep_embed}
\end{equation}
where $\gamma(\cdot)$ is a sinusoidal encoding and $\mathbf{y}_{\mathrm{txt}}$ represents the global text embedding. Within each transformer block, the affine parameters (scale $\alpha$, shift $\beta$, gate $g$) are dynamically generated based on the modality of the current token:
\begin{equation}
    \hat{\mathbf{z}}^{(i)} = \bigl(1 + \alpha_m^{(i)}\bigr) \cdot \mathrm{LN}(\mathbf{z}^{(i)}) + \beta_m^{(i)}, \quad m = \mathrm{mod}(i),
    \label{eq:adaln}
\end{equation}
where $\mathrm{mod}(i)$ identifies the modality of the $i$-th token. This mechanism allows the network to process different modalities at varying noise levels simultaneously, such as denoising the box layout while keeping the image tokens fully noisy.

\paragraph{Causal Attention Mask.}
To enforce a coarse-to-fine generation hierarchy, we impose a structured attention mask $\mathbf{A}$. We define a causal constraint wherein \textbf{layout queries are prohibited from attending to image keys}, while image tokens retain full access to box tokens. Formally:
\begin{equation}
    A_{ij} = \begin{cases}
        0 & \text{if } i \in \mathcal{S}_{\mathrm{box}} \text{ and } j \in \mathcal{S}_{\mathrm{img}}, \\
        1 & \text{otherwise}.
    \end{cases}
    \label{eq:mask}
\end{equation}
This prevents information leakage from the fine-grained image stream to the coarse layout stream, ensuring that the layout generation relies solely on textual and conditional semantics.

\subsection{Training Strategy}
\label{sec:training}

A core contribution of our framework lies in its unified multi-objective training strategy, which transcends standard image synthesis. Rather than training a static generator, we strategically decouple and couple the noise schedules of different modalities to simulate various conditional inference scenarios. This deliberately designed regime acts as the foundational enabler for CalliMaster's diverse downstream capabilities, unlocking not only coarse-to-fine generation but also layout-aware editing, artifact restoration, and multi-scale forensic identification. 

To optimize the multi-modal flow transformer, we define the per-modality loss as the mean squared error between the predicted velocity $\mathbf{v}_\theta$ and the target direction $\mathbf{u}_m = \boldsymbol{\epsilon}_m - \mathbf{m}_0$:
\begin{equation}
    \mathcal{L}_m(t_m) = \mathbb{E}_{t_m, \mathbf{m}_0, \boldsymbol{\epsilon}_m} \left[ \| \mathbf{v}_\theta(\mathbf{Z}_t, t_m, \dots) - (\boldsymbol{\epsilon}_m - \mathbf{m}_0) \|^2 \right],
    \label{eq:base_loss}
\end{equation}
where $m \in \{\mathrm{img, box, cond}\}$ represents the respective modality. Our framework employs a multi-objective training strategy by sampling one of four regimes at each iteration based on a probability distribution $\mathcal{P} = \{p_{\mathrm{S1}}, p_{\mathrm{S2}}, p_{\mathrm{joint}}, p_{\mathrm{uncond}}\}$:

    \paragraph{Stage 1: Box Generation.} The model learns structural planning. We set $t_{\mathrm{img}} = 1$ (pure noise), $t_{\mathrm{cond}} = 0$ (clean), and sample $t_{\mathrm{box}} \sim \sigma(\mathcal{N}(0,1))$, where $\sigma(\cdot)$ denotes the sigmoid function used to warp the normal distribution towards intermediate timesteps for better flow estimation. The objective prioritizes layout prediction:
    \begin{equation}
        \mathcal{L}_{\mathrm{S1}} = \mathcal{L}_{\mathrm{box}} + \lambda_{\mathrm{aux}}\,(\mathcal{L}_{\mathrm{img}} + \mathcal{L}_{\mathrm{cond}}).
        \label{eq:loss_s1}
    \end{equation}
    
    \paragraph{Stage 2: Content Synthesis.} The model learns content synthesis conditioned on the layout. We sample $t_{\mathrm{img}} \sim \sigma(\mathcal{N}(0,1))$ while fixing $t_{\mathrm{box}} = 0$ and $t_{\mathrm{cond}} = 0$. To enhance robustness, we inject a perturbation $\delta$ into the clean box latent:
    \begin{equation}
        \mathcal{L}_{\mathrm{S2}} = \mathcal{L}_{\mathrm{img}} + \lambda_{\mathrm{aux}}\,(\mathcal{L}_{\mathrm{box}} + \mathcal{L}_{\mathrm{cond}}).
        \label{eq:loss_s2}
    \end{equation}

    \paragraph{Joint Training (Inpainting).} All three visual modalities are corrupted with the \emph{same} noise level $t_{\mathrm{img}} = t_{\mathrm{box}} = t_{\mathrm{cond}} = t \sim \sigma(\mathcal{N}(0,1))$. While we observe that direct joint generation from pure noise yields suboptimal structural alignment compared to the cascaded approach, this regime is critical for \textbf{holistic inpainting}. The objective balances the reconstruction of all visual streams:
    \begin{equation}
        \mathcal{L}_{\mathrm{joint}} = \mathcal{L}_{\mathrm{img}} + \mathcal{L}_{\mathrm{box}} + \mathcal{L}_{\mathrm{cond}}.
        \label{eq:loss_joint}
    \end{equation}
    It enables the model to simultaneously recover missing layout structures and calligraphic content based on partial observations, serving as the foundation for our artifact restoration tasks.
    
    \paragraph{Unconditional Training.} To support classifier-free guidance and enhance the model's generative priors, we train the model to synthesize images without any external cues. We set $t_{\mathrm{img}} \sim \sigma(\mathcal{N}(0,1))$ while fixing the layout and condition timesteps to pure noise, i.e., $t_{\mathrm{box}} = 1$ and $t_{\mathrm{cond}} = 1$. The objective focuses solely on the image stream reconstruction:
    \begin{equation}
        \mathcal{L}_{\mathrm{uncond}} = \mathcal{L}_{\mathrm{img}} + \lambda_{\mathrm{aux}}\,(\mathcal{L}_{\mathrm{box}} + \mathcal{L}_{\mathrm{cond}}).
        \label{eq:loss_uncond}
    \end{equation}
    This regime forces the model to learn the intrinsic distribution of Chinese calligraphy styles and structures from the image data alone, significantly improving glyph stability and robustness~\cite{xu2025unicalli}.

\subsection{Inference and Downstream Applications}
\label{sec:inference_apps}

Our flow-matching framework unifies generation, editing, restoration, and forensics within a single, consistent mathematical formulation. We define the ODE integrator as $\Psi(\mathbf{z}_1, t_s \to t_e; \mathcal{C})$, a function that evolves a latent representation $\mathbf{z}$ from starting time $t_s$ to end time $t_e$ conditioned on the context $\mathcal{C}$.

\paragraph{Controllable Generation.}
Standard synthesis follows the cascaded protocol described in \cref{sec:training}. First, the layout is planned: $\hat{\mathbf{b}}_0 = \Psi(\mathbf{b}_1, 1 \to 0; \{\mathbf{c}, \text{txt}\})$. Second, the image is synthesized conditioned on this layout: $\hat{\mathbf{x}}_0 = \Psi(\mathbf{x}_1, 1 \to 0; \{\mathbf{c}, \hat{\mathbf{b}}_0, \text{txt}\})$.

\paragraph{Semantic Editing.}
For editing tasks (e.g., resizing characters or changing composition), users modify the bounding box layout $\mathbf{b}$ to $\mathbf{b}'$. We bypass the Stage~1 planner and directly execute the Stage~2 integrator:
\begin{equation}
    \hat{\mathbf{x}}'_{\text{edited}} = \Psi(\mathbf{x}_1, 1 \to 0; \{\mathbf{c}, \mathbf{b}', \text{txt}\}).
\end{equation}
By keeping the initial noise latent $\mathbf{x}_1$ fixed, the model inherently preserves the micro-level artistic identity (e.g., original brush textures and ink variations). Simultaneously, conditioned on the novel geometric prompt $\mathbf{b}'$, the generator is compelled to automatically re-infer the macro-level continuous brush momentum and inter-character ligatures. This dynamic adaptation aligns the redesigned composition with the updated geometric guidance, avoiding artifacts from simple pixel warping.

\paragraph{Artifact Restoration via Joint Inpainting.}
We treat restoration as a \textbf{constrained joint inference} problem. Given a damaged image $\mathbf{I}_{\text{corr}}$ and a binary mask $\mathbf{M}$ (where $M_{ij}=1$ denotes missing regions), we seek to recover the latent $\mathbf{z} = [\mathbf{x}, \mathbf{b}]$ in the missing regions while harmonizing with the known regions. We utilize the \textbf{Joint Training} regime ($t_{\text{img}}=t_{\text{box}}=t$), evolving both layout and image simultaneously.
At each integration step $t_i \to t_{i-1}$, we enforce consistency with the uncorrupted regions using the \emph{Replacement Method}~\cite{lugmayr2022repaint}:
\begin{equation}
    \mathbf{z}_{t_{i-1}} = \mathbf{M} \odot \hat{\mathbf{z}}_{t_{i-1}}^{\text{model}} + (1 - \mathbf{M}) \odot \underbrace{\bigl((1 - t_{i-1})\mathbf{z}_0^{\text{GT}} + t_{i-1}\boldsymbol{\epsilon}\bigr)}_{\text{Noisy Ground Truth}},
    \label{eq:inpainting}
\end{equation}
where $\hat{\mathbf{z}}_{t_{i-1}}^{\text{model}}$ is the ODE prediction and $\mathbf{z}_0^{\text{GT}}$ is the encoder projection of $\mathbf{I}_{\text{corr}}$. This allows CalliMaster to hallucinate missing strokes and layout structures solely within the masked area $\mathbf{M}$, ensuring seamless boundary transition. Experimental results and qualitative examples for this task are provided in Appendix~\ref{sec:app_restoration}.

\paragraph{Forensic Identification via Multi-Scale Reconstruction.}
We propose a metric, \emph{Diffusion Reconstruction Score} (DRS), to quantify stylistic authenticity. The core insight is that diffusion models act as frequency-selective filters: small noise levels ($t \to 0$) perturb high-frequency details (stroke edges, character glyphs), while large noise levels ($t \to 1$) disrupt low-frequency structures (overall layout). 
Given a query image $\mathbf{x}_{\text{query}}$, we construct a set of noise levels $\mathcal{T} = \{t_1, \dots, t_K\}$ spanning $[0, 1]$. For each $t_k$, we add noise to obtain $\mathbf{x}_{t_k}$ and perform a single-step flow reconstruction to estimate $\hat{\mathbf{x}}_{0}^{(t_k)}$:
\begin{equation}
    \hat{\mathbf{x}}_{0}^{(t_k)} = \mathbf{x}_{t_k} - t_k \cdot v_{\theta}(\mathbf{x}_{t_k}, t_k, \mathbf{c}, \mathbf{b}),
    \label{eq:reconstruction}
\end{equation}
where $v_{\theta}$ is the predicted velocity. The final authenticity score $\mathcal{S}$ is computed as the weighted ensemble of reconstruction errors:
\begin{equation}
    \mathcal{S}(\mathbf{x}_{\text{query}}) = \sum_{k=1}^{K} w_k \cdot \|\mathbf{x}_{\text{query}} - \hat{\mathbf{x}}_{0}^{(t_k)}\|_2^2.
    \label{eq:forensic_score}
\end{equation}
Authentic works, lying near the manifold center, exhibit low $\mathcal{S}$ across all scales. Forgeries typically manifest anomalies: high error at small $t$ indicates unnatural stroke artifacts (high-frequency), while high error at large $t$ indicates structural flaws (low-frequency).

\begin{figure}[!t]
    \centering
    \includegraphics[width=0.9\textwidth]{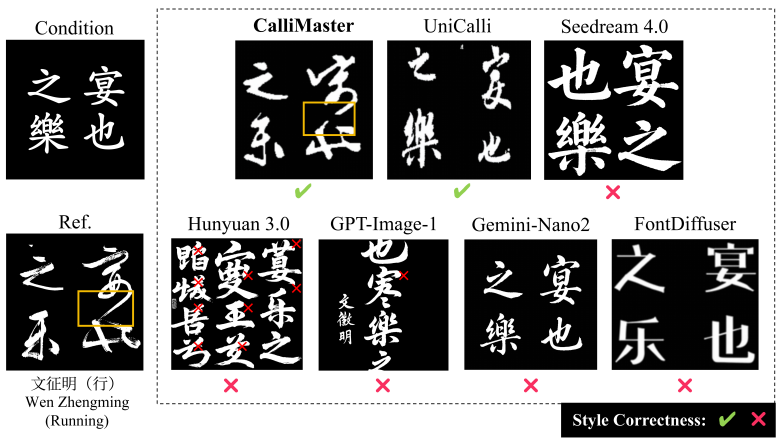} 
    \vspace{-1em}
    \caption{\textbf{Qualitative comparison of CalliMaster against state-of-the-art models.} A red cross \textcolor{red}{$\boldsymbol{\times}$} marks incorrectly rendered or hallucinated characters, and a red circle \textcolor{red}{$\bigcirc$} indicates omitted characters. Yellow boxes \textcolor{orange}{$\square$} highlight continuous strokes.}
    % \caption{\textbf{Qualitative comparison of CalliMaster against state-of-the-art models.} A red cross \textcolor{red}{$\boldsymbol{\times}$} marks incorrectly rendered or hallucinated characters, and a red circle \textcolor{red}{$\bigcirc$} indicates omitted characters. Yellow boxes \textcolor{orange}{$\square$} highlight well-formed continuous strokes.}
    \label{fig:comparison}
\end{figure}

\section{Experiments}

\begin{figure}[t]
    \centering
    \includegraphics[width=0.85\linewidth]{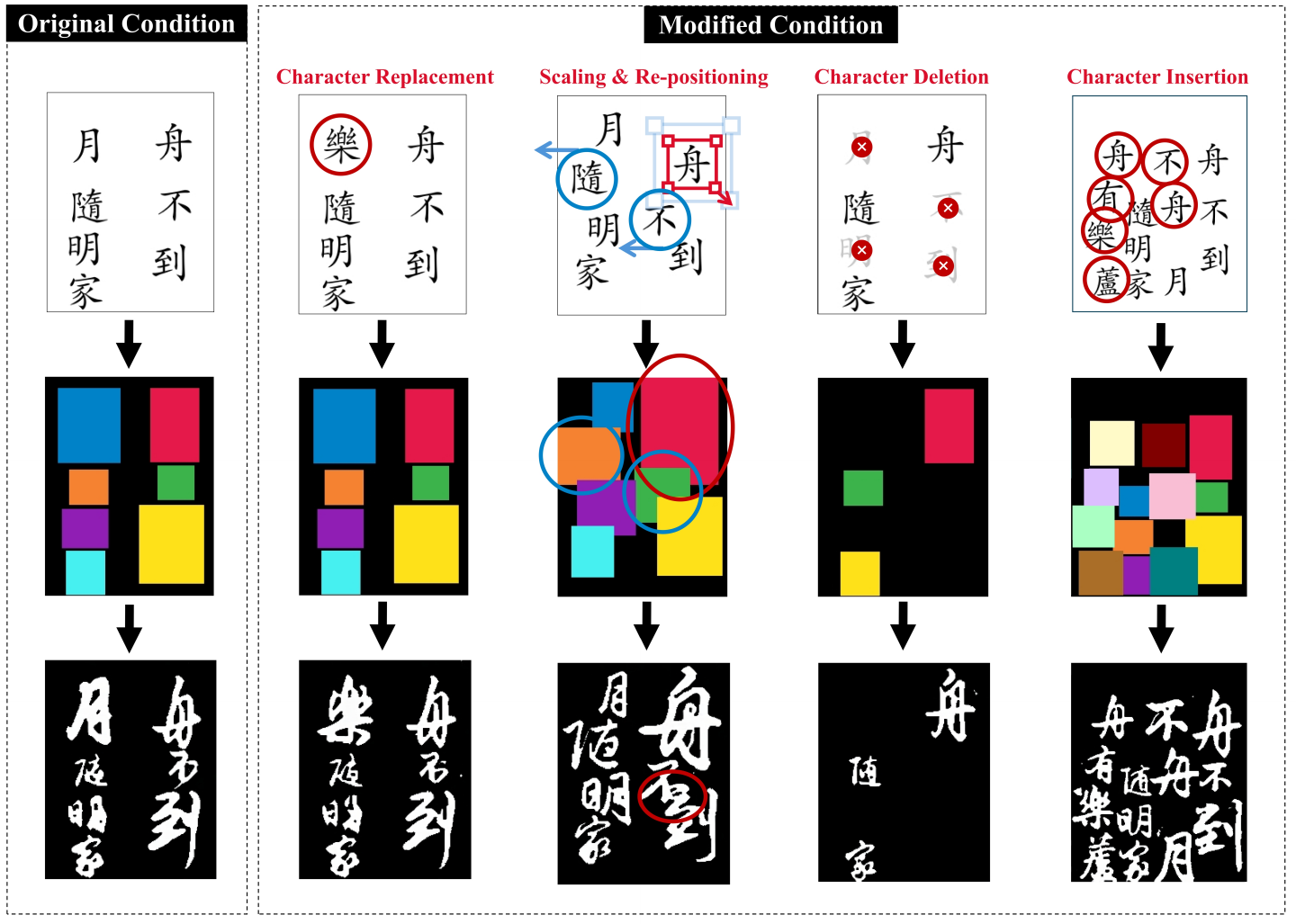}
    \captionof{figure}{\textbf{Semantic Layout Editing via Geometric Prompts.} CalliMaster supports diverse interactive operations (replacement, scaling, deletion, and insertion). By manipulating bounding boxes, the model re-harmonizes continuous strokes and adjusts the spatial composition to maintain visual rhythm.}
\label{fig:semantic_editing}
\end{figure}

\begin{table}[t]
\centering
\caption{\textbf{Quantitative comparison of calligraphy generation methods.} CLIP/GPT scores measure stylistic similarity to the reference. CharsAcc is the F1 score from PP-OCR~\cite{cui2025paddleocr30technicalreport}.}
\label{tab:comparison}
\scalebox{0.8}{
\begin{tabular}{lccc}
\toprule
Method & CLIP-Score$\uparrow$ & GPT-Score$\uparrow$ & CharsAcc$\uparrow$ \\
\midrule
FontDiffuser & 0.8029 & 2.4 & \textbf{75.52\%} \\
Gemini-Nano2 & 0.7620 & 2.2 & 69.23\% \\
Hunyuan 3.0 & 0.7547 & 2.3 & 38.47\% \\
Seedream 4.0 & 0.7864 & 2.6 & 62.54\% \\
GPT-Image-1 & 0.7897 & 2.7 & 50.18\% \\
UniCalli & 0.9051 & 3.6 & 42.03\% \\
\midrule
\textbf{CalliMaster} (pred. box) & 0.9574 & 3.9 & 33.41\% \\
\textbf{CalliMaster} (given box) & \textbf{0.9663} & \textbf{4.1} & 33.56\% \\
\midrule
\color{gray}{GT (Reference)} & \color{gray}{--} & \color{gray}{--} & \color{gray}{34.12\%} \\
\bottomrule
\end{tabular}}
\end{table}

\subsection{Implementation Details}
\label{sec:implementation}
We build CalliMaster on the Flux architecture~\cite{labs2025flux1kontextflowmatching} ($d_{\mathrm{model}}{=}3072$, 19 double-stream + 38 single-stream blocks) with a frozen T5-XXL~\cite{raffel2023exploring} encoder. Training is conducted on 8$\times$H100 GPUs with batch size 8, using AdamW ($\text{lr}{=}1{\times}10^{-5}$). We set regime probabilities $p_{\mathrm{S1}}{=}0.35$, $p_{\mathrm{S2}}{=}0.35$, $p_{\mathrm{joint}}{=}0.25$, $p_{\mathrm{uncond}}{=}0.05$, with $\lambda_{\mathrm{aux}}{=}0.01$ and box perturbation $\delta \sim \mathcal{U}(0, 0.1)$. Each sample contains up to 20 characters at a canonical font size of 64\,px. We use a 25-step rectified flow ODE solver at inference. Data preparation details are in Appendix~\ref{sec:app_data}.

\subsection{Comparison with State-of-the-Art Methods}
\label{sec:comparison}

We compare CalliMaster against specialized calligraphy models (UniCalli~\cite{xu2025unicalli}, FontDiffuser~\cite{yang2024fontdiffuser}) and general-purpose foundational models (GPT-Image-1~\cite{openai2025gptimage}, Hunyuan 3.0~\cite{cao2025hunyuanimage}, Seedream 4.0~\cite{seedream4}, and Gemini-Nano-Banana2~\cite{google2026nanobanana2}). All baseline prompts were optimized for fair comparison (details in Suppl.).

\noindent\textbf{Quantitative Evaluation.} 
As shown in \cref{tab:comparison}, CalliMaster achieves new state-of-the-art CLIP-Score ($0.9663$) and GPT-Score ($4.1$). A notable observation concerns CharsAcc: while FontDiffuser scores highest ($75.52\%$), it generates rigid, font-like glyphs. Authentic cursive calligraphy is inherently challenging for OCR---the GT reference itself only yields $34.12\%$. CalliMaster's CharsAcc ($33.56\%$) closely matches the GT distribution, indicating it captures artistic complexity instead of producing standardized fonts.

\noindent\textbf{Qualitative Evaluation.} 
As shown in \cref{fig:comparison}, baselines frequently suffer from character omissions or structural hallucinations, while CalliMaster consistently produces high-quality page-scale calligraphy with harmonious spatial rhythms and well-formed continuous strokes.

\begin{table}[b]
  \centering
  \small
  \caption{
      \textbf{User study results} (5-point Likert scale, mean$\pm$std). Best scores in \textbf{bold}.
  }
  \label{tab:user_study}
  \vspace{-.1 in}
  \scalebox{0.75}{
  \begin{tabular}{@{}lccccc@{}}
  \toprule
  \textbf{Method} & \textbf{Style Fidelity$\uparrow$} & \textbf{Glyph Accuracy$\uparrow$} & \textbf{Naturalness$\uparrow$} & \textbf{Layout Similarity$\uparrow$} & \textbf{Overall$\uparrow$} \\
  \midrule
  FontDiffuser               & 1.720, 1.385 & \textbf{4.920, 0.396} & 2.080, 1.512 & 1.493, 1.287 & 1.840, 1.346 \\
  GPT-Image-1                & 3.147, 1.178 & 3.720, 1.215 & 2.587, 1.302 & 2.360, 1.194 & 2.680, 1.253 \\
  Hunyuan 3.0                    & 2.653, 1.268 & 4.107, 0.984 & 3.187, 1.245 & 2.573, 1.136 & 3.013, 1.108 \\
  Gemini-Nano2               & 2.373, 1.312 & 3.547, 1.278 & 2.840, 1.356 & 2.187, 1.224 & 2.453, 1.285 \\
  UniCalli                   & 4.187, 0.968 & 4.760, 0.478 & 4.413, 0.782 & 4.080, 0.914 & 4.467, 0.812 \\
  \midrule
  \textbf{CalliMaster (Ours)} & \textbf{4.427, 0.856} & 4.853, 0.384 & \textbf{4.613, 0.724} & \textbf{4.520, 0.738} & \textbf{4.680, 0.674} \\
  \bottomrule
  \end{tabular}}
  \vspace{-.1 in}
\end{table}

\noindent\textbf{User Study.} 
We further conduct a human evaluation on a 5-point Likert scale across five perceptual dimensions (\cref{tab:user_study}). CalliMaster achieves the highest overall score ($4.68$) and leads in style fidelity, naturalness, and layout similarity, while maintaining near-best glyph accuracy ($4.85$ vs.\ FontDiffuser's $4.92$). The low standard deviations indicate strong inter-rater agreement on CalliMaster's superior quality.

\begin{table}[t]
\centering
\caption{\textbf{Diffusion Reconstruction Score (DRS) across methods.} ``GT+Noise'' denotes GT with blob artifacts at increasing intensities (L/M: 8/20 blobs).}
\label{tab:drs}\scalebox{0.75}{
\begin{tabular}{lcccccccc}
\toprule
Method & GT & GT+Noise(L) & GT+Noise(M) & CalliMaster & FontDiffuser & Gemini-Nano2 & GPT-Image-1 & Seedream 4.0 \\
\midrule
DRS $\downarrow$ & \textbf{0.186} & 0.237 & 0.472 & 0.299 & 0.561 & 0.826 & 0.962 & 0.997 \\
\bottomrule
\end{tabular}}
\end{table}

\begin{figure}[t]
    \centering
    \includegraphics[width=\linewidth]{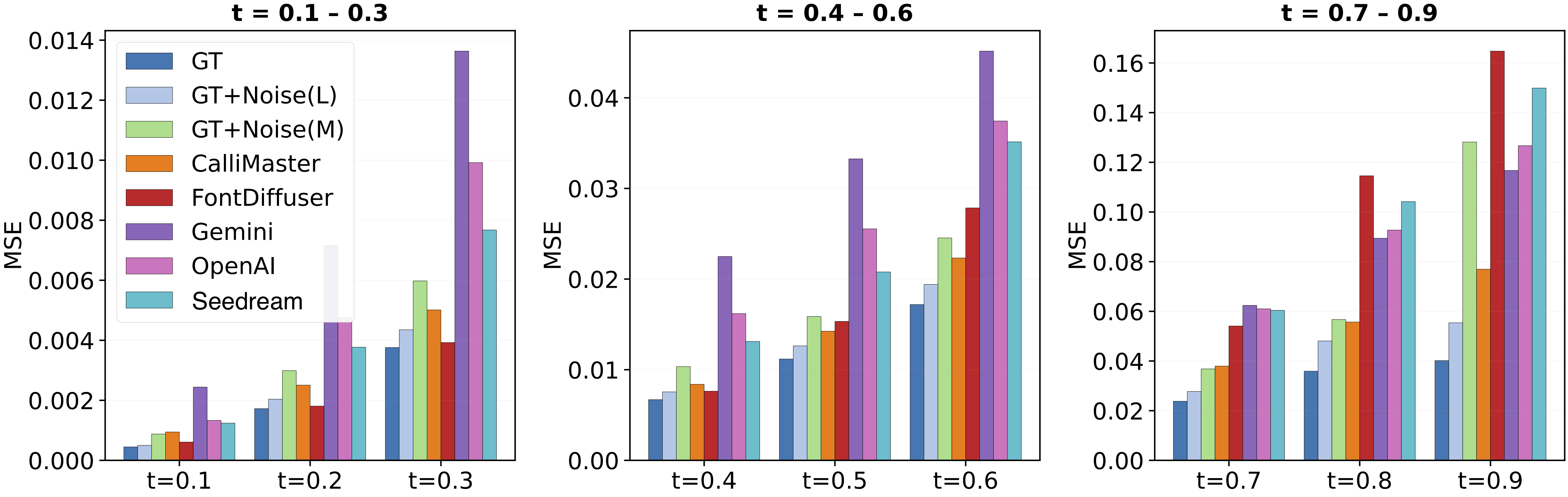}
    \caption{\textbf{Per-timestep DRS at each noise scale.} Low $t$ probes stroke-level fidelity; high $t$ probes global layout plausibility.}
    \label{fig:drs}
\end{figure}

\subsection{Semantic Re-planning and Editing}
\label{sec:semantic_editing_exp}

Beyond static generation, CalliMaster demonstrates robust capabilities in semantic layout editing by treating intermediate bounding boxes as explicit geometric prompts. As illustrated in \cref{fig:semantic_editing}, the model dynamically re-harmonizes the entire page in response to any layout-level perturbation. Specifically, it re-infers inter-character ligatures and continuous brush trajectories to bridge new spatial relationships, while redistributing the negative space to maintain the authentic rhythmic flow essential to calligraphy.

\subsection{Ablation Study}
\label{sec:ablation}

\noindent\textbf{Key Components.} As shown in \cref{tab:ablation_components}, removing box prediction degrades CLIP-Score from $0.9234$ to $0.8698$ and LPIPS from $0.1768$ to $0.2149$, confirming that decoupling spatial planning from content synthesis is essential. Removing the modulate embedding causes an even larger drop (CLIP-Score $0.7400$, LPIPS $0.3782$), underscoring its critical role in capturing style-specific brushwork.

\begin{figure}[t]
    \centering
    \includegraphics[width=0.9\linewidth]{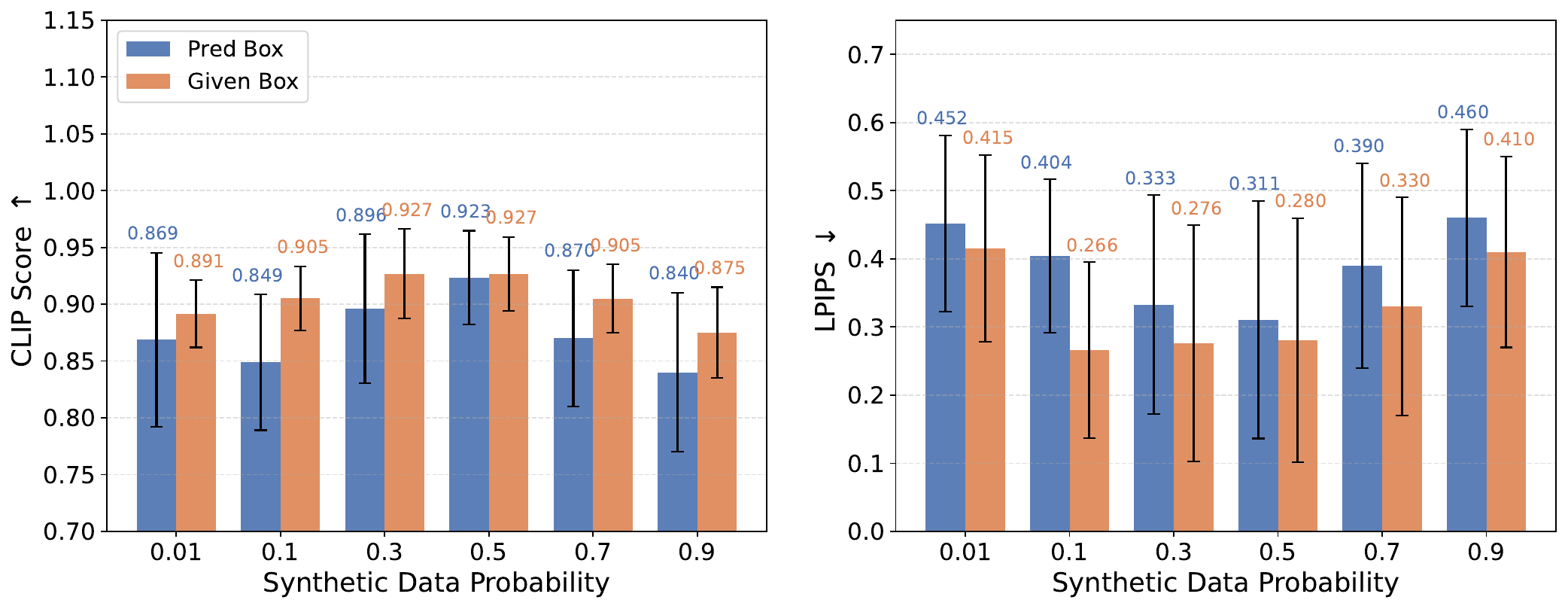}
    \caption{\textbf{Effect of Synthetic Data Proportion.} We analyze the impact of the synthetic data sampling probability ($p_{\mathrm{synth}}$) on generation quality. The bar charts illustrate the CLIP Score (left) and LPIPS (right) across different training ratios, evaluated under both predicted and given bounding box settings. }
    \label{fig:ablation_synth}
\end{figure}

\noindent\textbf{Synthetic Data Proportion.} \cref{fig:ablation_synth} shows that too little synthetic data limits layout generalization (low CLIP-Score with predicted boxes), while too much erodes authentic brush textures (worse LPIPS). The optimal balance is achieved at $p_{\mathrm{synth}} {=} 0.5$.

\subsection{Calligraphy Forensic Identification}
We evaluate the Diffusion Reconstruction Score (DRS) for forensic identification. DRS measures single-step reconstruction MSE across noise scales; images matching the model's learned distribution yield lower scores. \cref{tab:drs} shows DRS increases monotonically with GT perturbation severity ($0.186 {\to} 0.237 {\to} 0.472$). CalliMaster achieves the lowest DRS ($0.299$) among all generation methods. \cref{fig:drs} reveals the score gap against general-purpose models widens at mid-to-high noise scales, indicating they fail to capture glyph-level structures and global layout plausibility (details in Appendix~\ref{sec:app_drs}).

\begin{table}[t]
    \centering
    \caption{\textbf{Ablation Study on Model Components.} Quantitative evaluation of key components in our framework.}
    \label{tab:ablation_components}
    \scalebox{0.8}{
    \begin{tabular}{lcc}
    \toprule
    \textbf{Method} & ~CLIP-Score$\uparrow$ & ~LPIPS$\downarrow$ \\
    \midrule
    \textbf{CalliMaster} (Ours) & \textbf{0.9234} & \textbf{0.1768} \\
    w/o Box Prediction          & 0.8698          & 0.2149       \\
    w/o Modulate Embedding      & 0.7400          & 0.3782      \\
    \bottomrule
    \end{tabular}}
\end{table}

\section{Conclusion}
\label{sec:conclusion}

We presented \textbf{CalliMaster}, a Multimodal Diffusion Transformer that decouples page-level Chinese calligraphy synthesis into a coarse-to-fine pipeline (Text $\rightarrow$ Layout $\rightarrow$ Image). This approach achieves both high-fidelity local brushwork and globally coherent layouts. Furthermore, intermediate layouts act as geometric prompts to enable user-guided semantic re-planning with dynamically regenerated ligatures. Beyond generation, CalliMaster supports layout-aware restoration and forensic identification via our Diffusion Reconstruction Score (DRS), achieving state-of-the-art performance across all evaluated tasks.

\bibliographystyle{splncs04}
\bibliography{main}

% ---- Appendix ----
\clearpage
\input{X_supp}

\end{document}

%% file: X_supp.tex
\appendix
\section{Data Preparation}
\label{sec:app_data}

\paragraph{Realistic Data.}
We integrate the public UniCalli dataset~\cite{xu2025unicalli} with a newly collected dataset of authentic calligraphy works sourced from the internet, spanning diverse scripts (Kai/Regular, Xing/Running, Cao/Cursive) and historical periods. For the newly collected images, we perform manual annotation to provide per-character bounding boxes and Unicode labels.
Given that raw web data frequently suffers from background clutter and paper degradation, we implement a specialized U-Net based denoising model to enhance stroke clarity. Following the training paradigm of ESRGAN~\cite{wang2018esrgan}, we optimize the model using adversarial and perceptual losses. The training data is synthesized by injecting random degradation artifacts into clean samples. Characters are clustered into columns, filtered for spatial continuity, and cropped into page-level samples. The condition image $\mathbf{c}$ is rendered using standard fonts, while the layout $\mathbf{b}$ encodes character regions with random colors.

\paragraph{Synthetic Data.}
Following UniCalli~\cite{xu2025unicalli}, we construct a synthetic data pipeline. For each sample, we randomly select $k \in [3, 20]$ characters and generate layouts using four modes: \emph{grid}, \emph{column}, \emph{random}, and \emph{scatter}. We apply per-character scale jitter $s \sim \mathcal{U}(0.85, 1.15)$ and aspect ratio perturbation $r \sim \mathcal{U}(0.9, 1.1)$. High-quality calligraphy TTF fonts are rendered into these boxes, yielding paired triplets (image, condition, box) with pixel-perfect alignment.

\paragraph{Merging Strategy and Robustness.}
Each batch element is drawn from either the realistic or synthetic pool via Bernoulli sampling with probability $p_{\mathrm{synth}}$. Synthetic data serves a dual role: (1)~it provides pixel-perfect box--image alignment that teaches the model precise layout-to-content mapping, compensating for annotation noise in realistic data; and (2)~it simulates the editing scenario. Because synthetic layouts are procedurally generated with arbitrary spatial perturbations, the model learns to render plausible calligraphy for any box configuration. This exposure to diverse, non-naturalistic layouts prevents overfitting to the spatial priors of the realistic corpus and ensures generalization to user-specified edits at inference time.

\section{Forensic Identification: DRS Evaluation Protocol}
\label{sec:app_drs}

We evaluate the Diffusion Reconstruction Score (DRS) at nine uniformly spaced noise levels ($t = 0.1$ to $0.9$), which capture different aspects of calligraphic authenticity:
\begin{itemize}
    \item \textbf{Low noise scales} ($t = 0.1$--$0.3$): Only fine-grained high-frequency details are corrupted, so reconstruction error reflects subtle stroke quality differences---brush pressure, stroke connections, and ink texture.
    \item \textbf{Mid noise scales} ($t = 0.4$--$0.6$): Character-level structure is disrupted, and reconstruction error captures the fidelity of individual glyph shapes.
    \item \textbf{High noise scales} ($t = 0.7$--$0.9$): The global spatial layout is destroyed, and reconstruction error measures the plausibility of the overall composition.
\end{itemize}

Since no ground-truth authentic--counterfeit pairs exist for Chinese calligraphy, we evaluate DRS by comparing outputs from different generation methods and by applying controlled perturbations (randomly placed elliptical blob artifacts at intensities L: 8 blobs, M: 20 blobs, H: 40 blobs) to real calligraphy. DRS is computed as the cumulative single-step reconstruction MSE over 9 noise levels with 3 trials each. See \cref{tab:drs} and \cref{fig:drs} in the main paper for the quantitative results and per-timestep visualization. 

\section{Artifact Restoration}
\label{sec:app_restoration}
\begin{figure}[H]
\vspace{-1em}
\centering
\begin{minipage}[t]{0.48\textwidth}
    \vspace{1pt}
    \centering
    \caption{Inpainting example.}
    \vspace{-6pt}
    \includegraphics[width=0.6\linewidth]{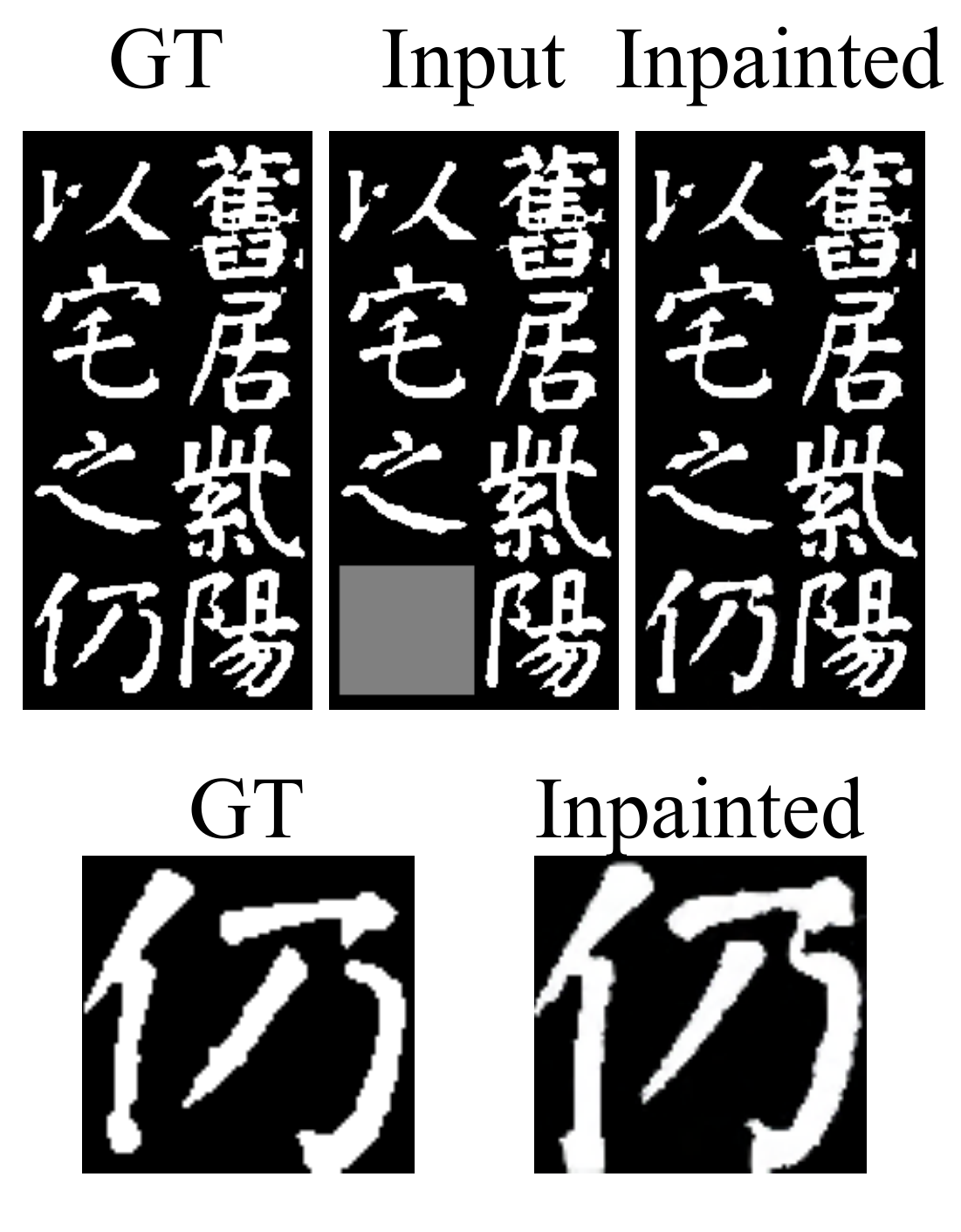}
    \label{fig:inpaint_example}
\end{minipage}
\hfill
\begin{minipage}[t]{0.48\textwidth}
    \vspace{-10pt}
    \centering
    \captionof{table}{Inpainting quality on 10 masked regions from real calligraphy images.}
    \label{tab:inpaint_eval}
    \vspace{2pt}
    \scalebox{0.8}{
    \begin{tabular}{ccc}
    \toprule
    Sample & LPIPS $\downarrow$ & CLIP Score $\uparrow$ \\
    \midrule
    0 & 0.0942 & 0.9650 \\
    1 & 0.1758 & 0.9169 \\
    2 & 0.1460 & 0.9709 \\
    3 & 0.0330 & 0.9806 \\
    4 & 0.1461 & 0.9381 \\
    5 & 0.0998 & 0.9689 \\
    6 & 0.2485 & 0.8295 \\
    7 & 0.0525 & 0.9842 \\
    8 & 0.0651 & 0.9871 \\
    9 & 0.0918 & 0.9822 \\
    \midrule
    \textbf{Mean} & \textbf{0.1153} & \textbf{0.9523} \\
    \bottomrule
    \end{tabular}}
\end{minipage}
\end{figure}

Due to the lack of paired real-world datasets for calligraphy restoration, we simulate realistic damage by masking character regions from pristine calligraphy images. We evaluate CalliMaster's joint inpainting capability on 10 such masked samples. As shown in \cref{fig:inpaint_example} and \cref{tab:inpaint_eval}, the model achieves a mean LPIPS of 0.1153 and CLIP Score of 0.9523, demonstrating faithful reconstruction of missing strokes while preserving spatial coherence with the surrounding context.

\section{User Study Details}
\label{sec:app_user_study}

\begin{figure}[H]
    \centering
    \includegraphics[width=\linewidth]{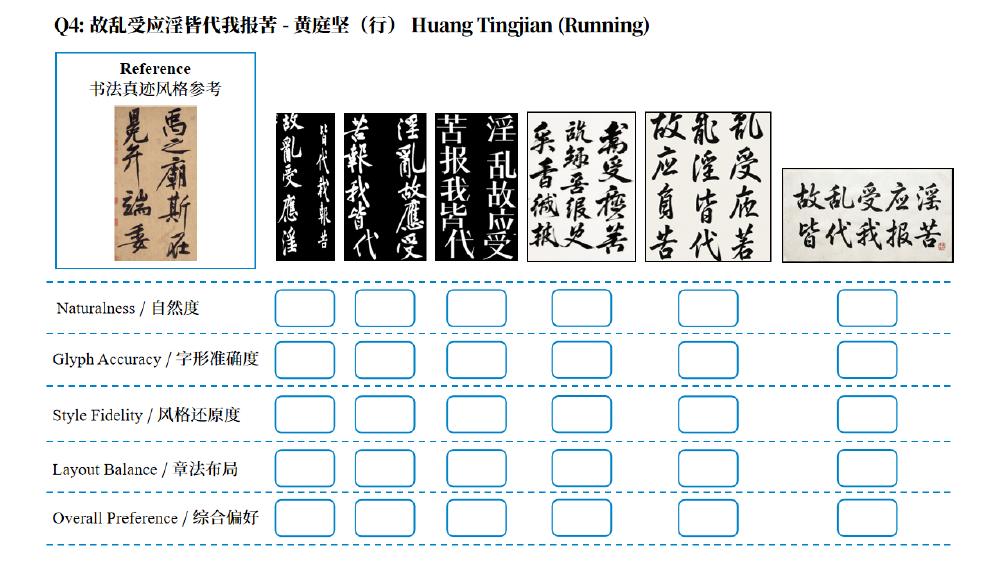}
    \caption{\textbf{Sample user study questionnaire for calligrapher Huang Tingjian.} This image displays one specific evaluation case. It presents 6 randomly ordered outputs from the compared methods. Evaluators rate each on Style Fidelity, Glyph Accuracy, Naturalness, Layout Similarity, and Overall Quality using a 5-point Likert scale.}
    \label{fig:user_study_htj}
\end{figure}

\begin{figure}[H]
    \centering
    \includegraphics[width=\linewidth]{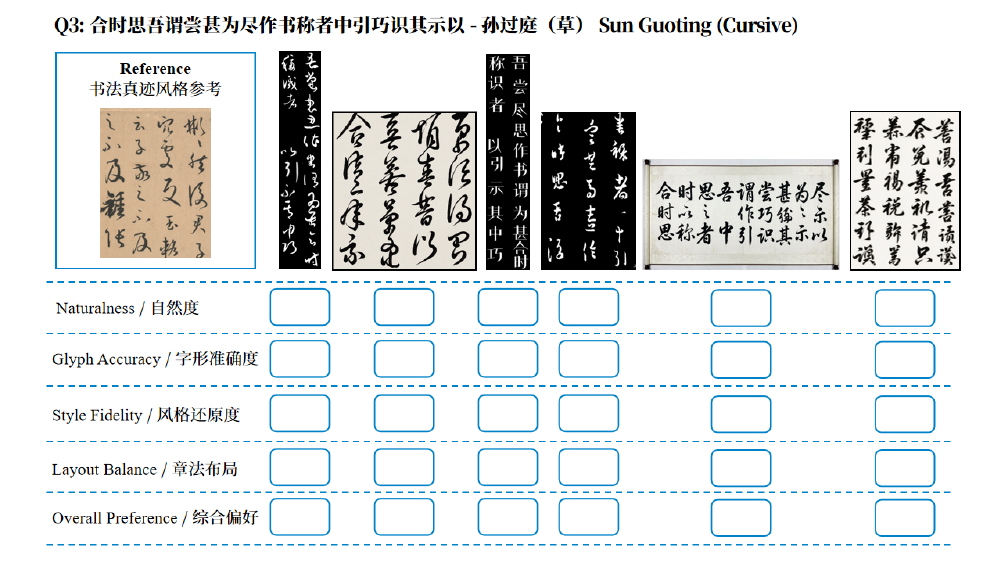}
    \caption{\textbf{Sample user study questionnaire for calligrapher Sun Guoting.} The same 10-case, 6-method, 5-criterion protocol as in \cref{fig:user_study_htj} is applied, targeting a distinct calligraphic style.}
    \label{fig:user_study_sgt}
\end{figure}

We conducted a structured perceptual evaluation comprising 10 cases from two target calligraphers (Huang Tingjian and Sun Guoting). Each case presents outputs from 6 compared methods---FontDiffuser, GPT-Image-1, Hunyuan~3.0, Gemini-Nano2, UniCalli, and CalliMaster (Ours)---shuffled in random order and presented without method labels to prevent position bias and evaluator bias. Twenty evaluators, including calligraphy enthusiasts and practicing professionals, rated each image on five dimensions (Style Fidelity, Glyph Accuracy, Naturalness, Layout Similarity, and Overall Quality) via a 5-point Likert scale. Quantitative results are reported in \cref{tab:user_study}; sample questionnaire interfaces are shown in \cref{fig:user_study_htj,fig:user_study_sgt}.

\vspace{-4em}

% -------------------------------------------------------
\section{Semantic Re-planning and Editing}
\label{sec:app_semantic_editing}

\begin{figure}[H]
\vspace{-1em}
    \centering
    \includegraphics[width=0.95\linewidth]{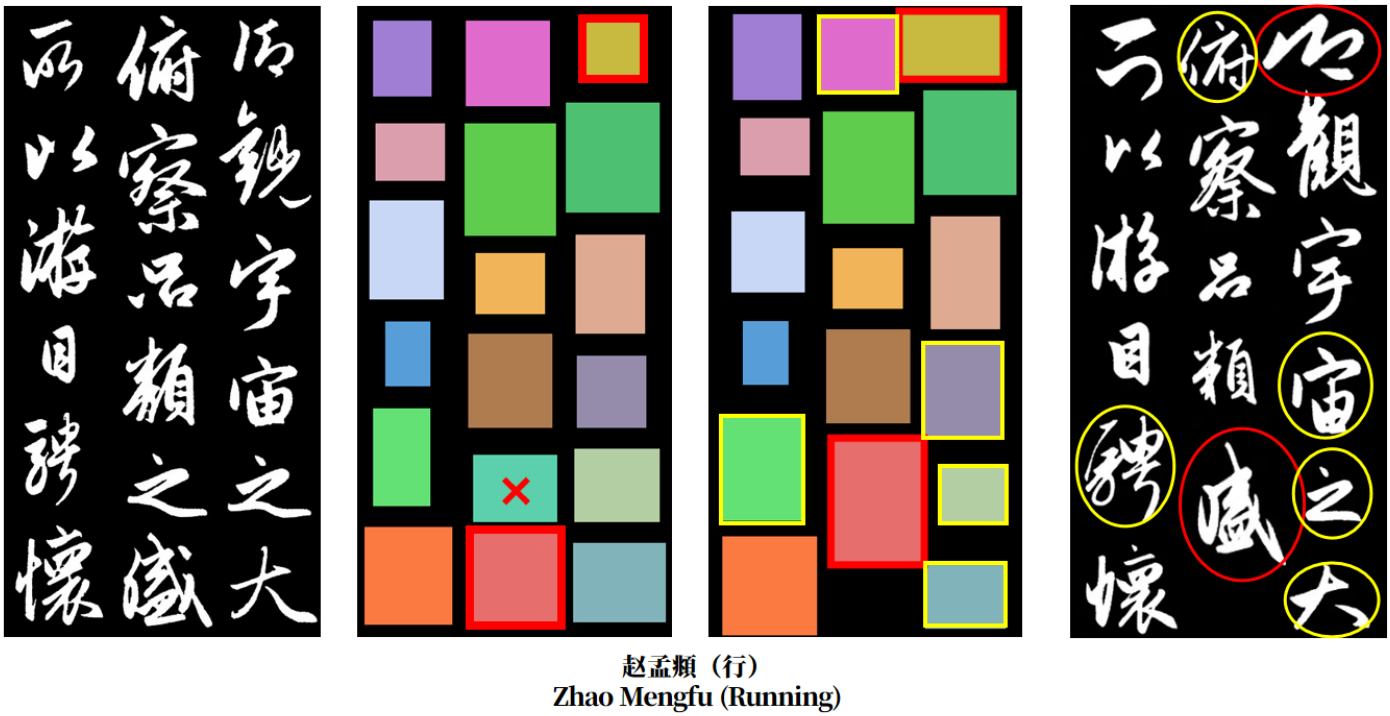}
    \includegraphics[width=0.95\linewidth]{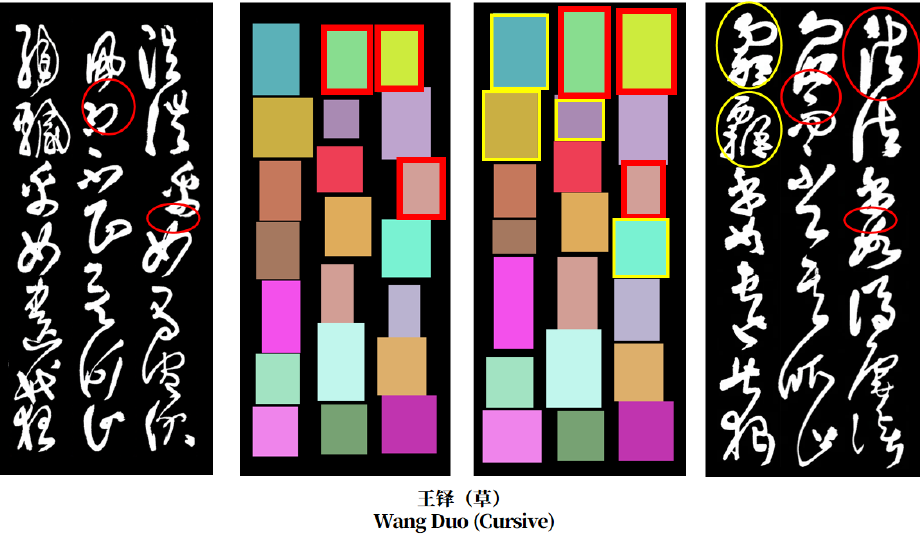}
    \caption{\textbf{Semantic editing cases.} The two left images display the initially generated calligraphy and its corresponding bounding box layout. Red outlines \textcolor{red}{$\square$} and cross \textcolor{red}{$\boldsymbol{\times}$} marks denote user edits including box enlargement, repositioning, and deletion. Following these modifications, the system regenerates the image with adapted character structures and overall layout. The model successfully creates natural stroke connections between characters. Red circles \textcolor{red}{$\bigcirc$} indicate the directly edited characters. Yellow markers highlight the surrounding text layouts that dynamically adjust to the new spatial composition.}
    \label{fig:semantic_editing}
\end{figure}

Beyond static generation, CalliMaster demonstrates robust capabilities in semantic layout editing by treating intermediate bounding boxes as explicit geometric prompts. As illustrated in \cref{fig:semantic_editing}, the model dynamically re-harmonizes the entire page in response to any layout-level perturbation.
% Specifically, it re-infers inter-character ligatures and continuous brush trajectories to bridge new spatial relationships, while redistributing the negative space to maintain the authentic rhythmic flow essential to calligraphy.

% -------------------------------------------------------
\section{Style-Conditioned Spatial Layout and Generation}
\label{sec:app_box}

\begin{figure}[H]
    \centering
    \includegraphics[width=0.9\linewidth]{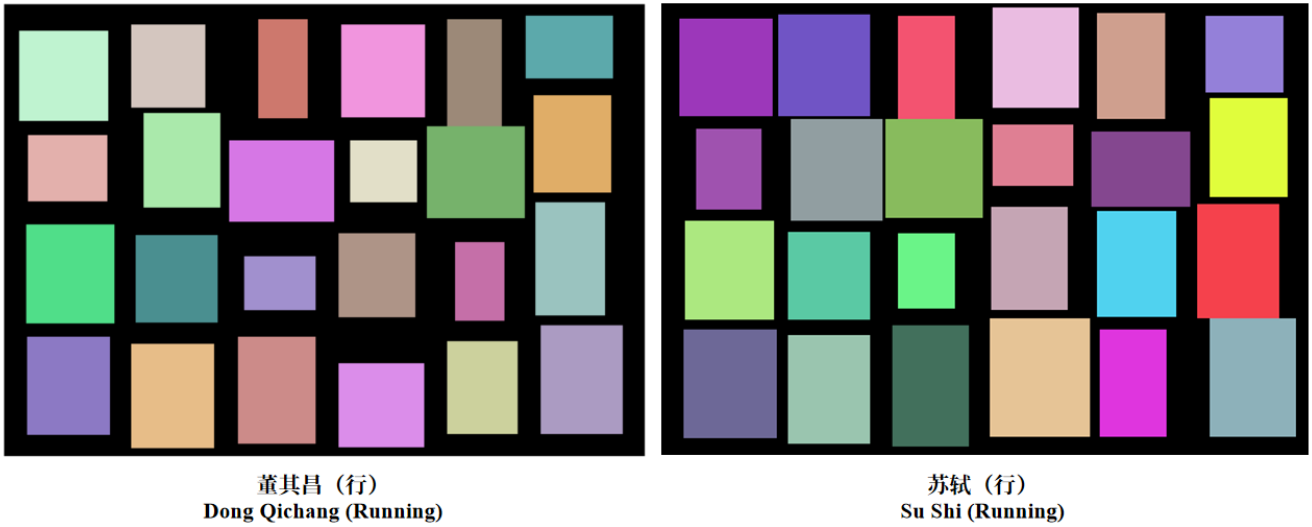}
    \includegraphics[width=0.9\linewidth]{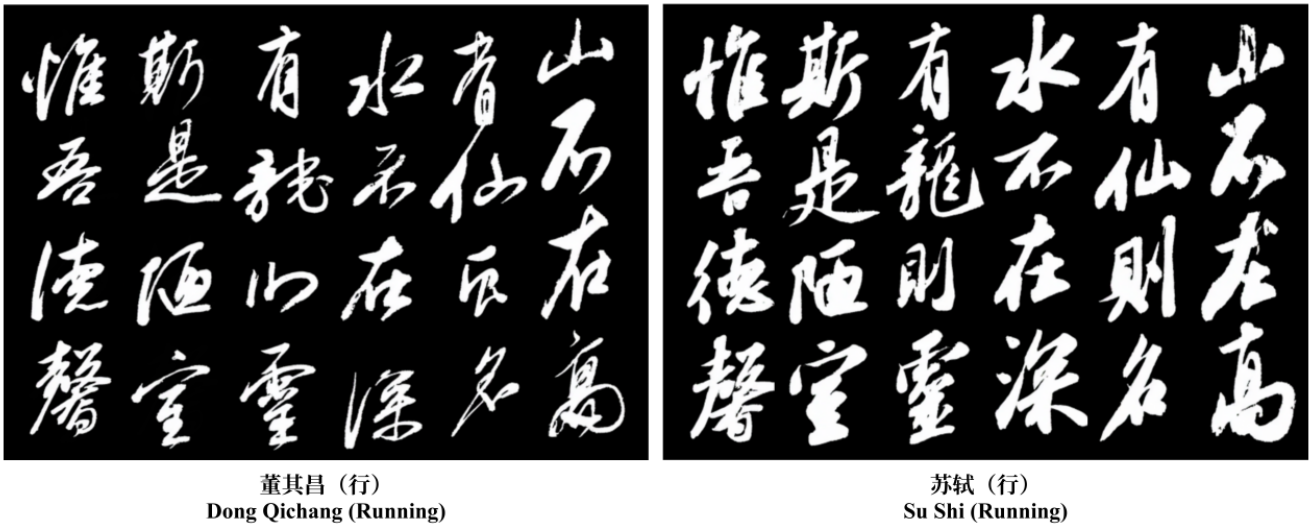}
    \caption{\textbf{Style-conditioned results for the first group.} \textbf{Top:} spatial bounding-box layouts generated for calligraphers' styles, showing the inferred character sizes, positions, and column arrangement. \textbf{Bottom:} calligraphy images generated from the layout above, exhibiting the brushwork and stroke character distinctive to these calligraphers.}
    \label{fig:box_p1}
\end{figure}

\begin{figure}[H]
    \centering
    \includegraphics[width=\linewidth]{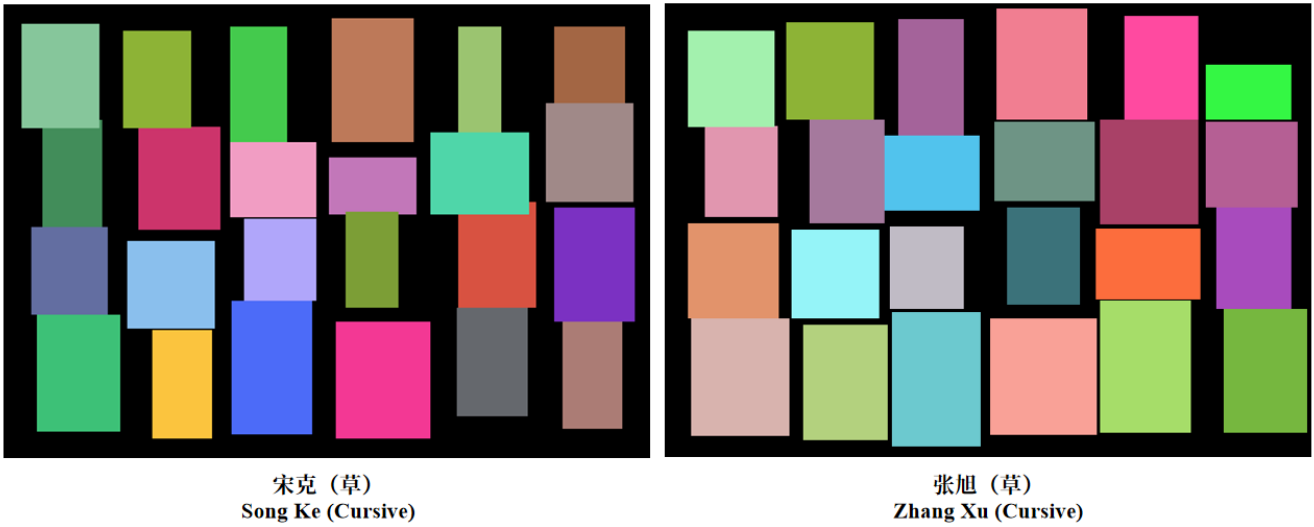}
    \includegraphics[width=\linewidth]{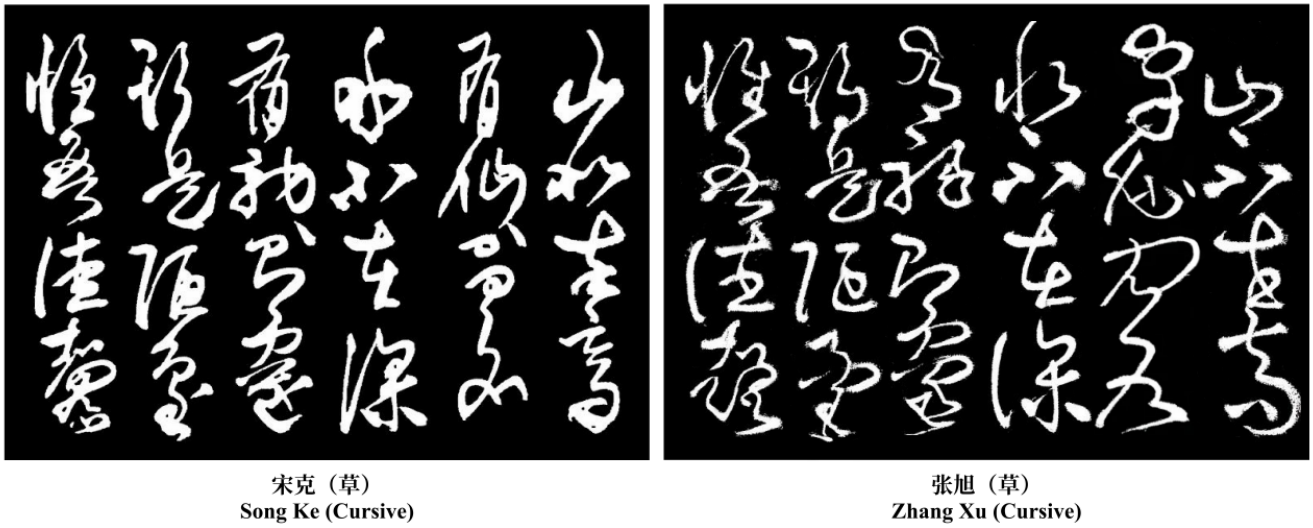}
    \caption{\textbf{Style-conditioned results for the second group.} \textbf{Top:} spatial layout adapted to the target styles under the exact same text condition. Note the distinct spatial rhythm and character proportions compared with group 1. \textbf{Bottom:} generated calligraphy reflecting these calligraphers' individual aesthetics.}
    \label{fig:box_p2}
\end{figure}

\begin{figure}[H]
    \centering
    \includegraphics[width=\linewidth]{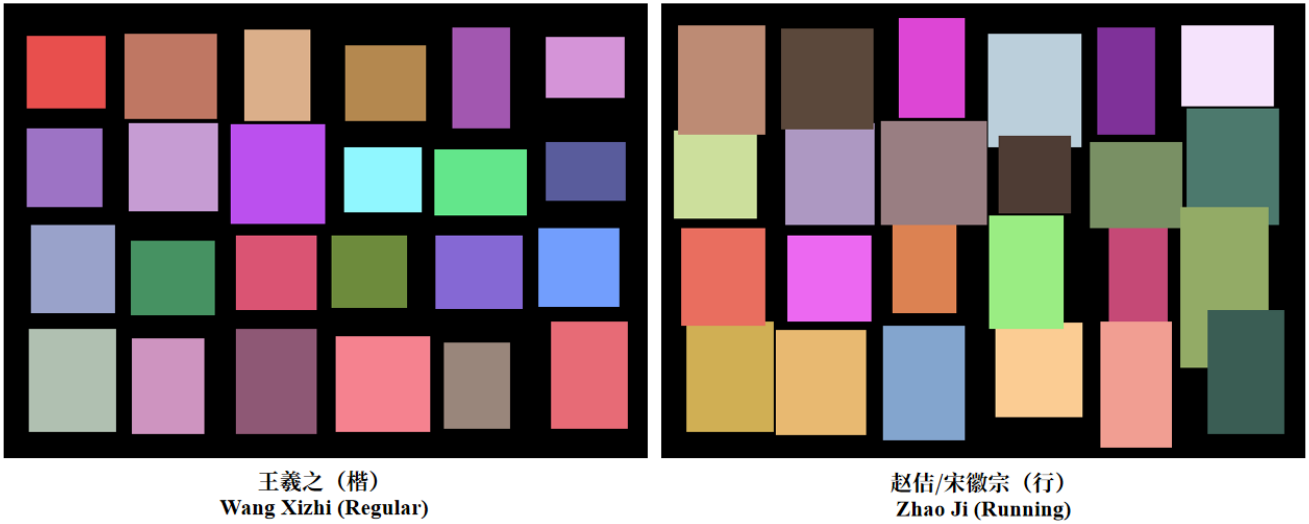}
    \includegraphics[width=\linewidth]{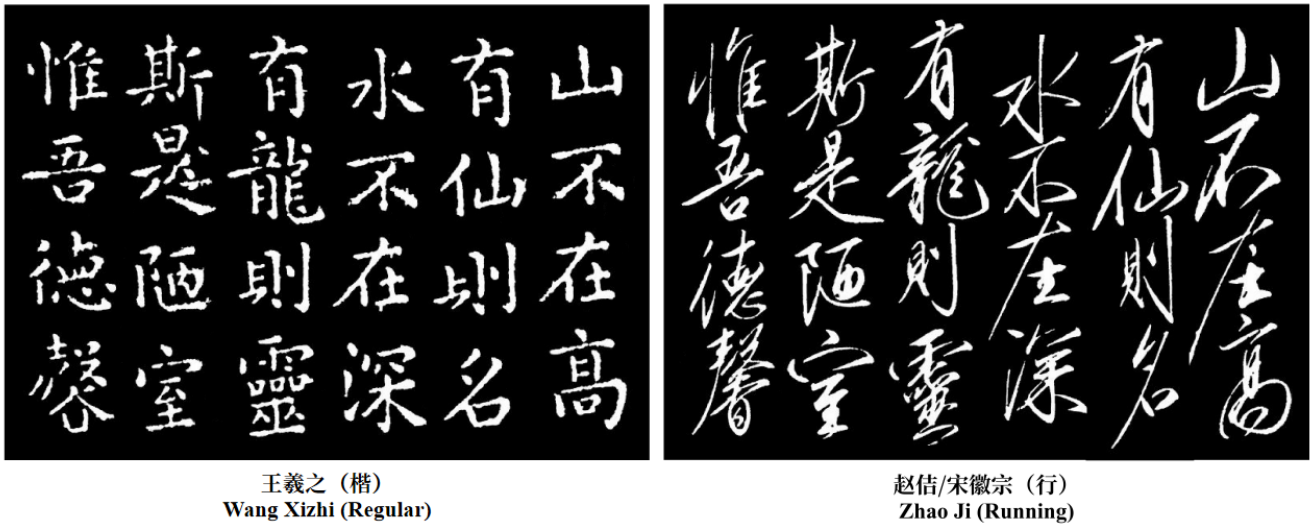}
    \caption{\textbf{Style-conditioned results for the third group.} \textbf{Top:} spatial layout for calligraphers' styles, demonstrating yet another spatial composition strategy induced by calligraphic style conditioning. \textbf{Bottom:} the corresponding generated images, illustrating how CalliMaster faithfully renders style-specific stroke dynamics and page-level composition.}
    \label{fig:box_p3}
\end{figure}

Figures~\ref{fig:box_p1}--\ref{fig:box_p3} present three groups of results, each sharing the same text condition but targeting a different calligrapher and script style. For each group, the \emph{top} image shows the spatial bounding-box layout inferred by CalliMaster for that particular calligraphic style, and the \emph{bottom} image shows the corresponding generated calligraphy output. These results demonstrate that CalliMaster adapts its brushwork and stroke style to match the individual characteristics of each target calligrapher. The system also adapts its spatial composition strategy by adjusting character sizing, inter-character spacing, and column structure.
%洗澡去了

\section{Additional Generated Samples}
\label{sec:app_samples}

\begin{figure}[H]
    \centering
    \includegraphics[width=0.9\linewidth]{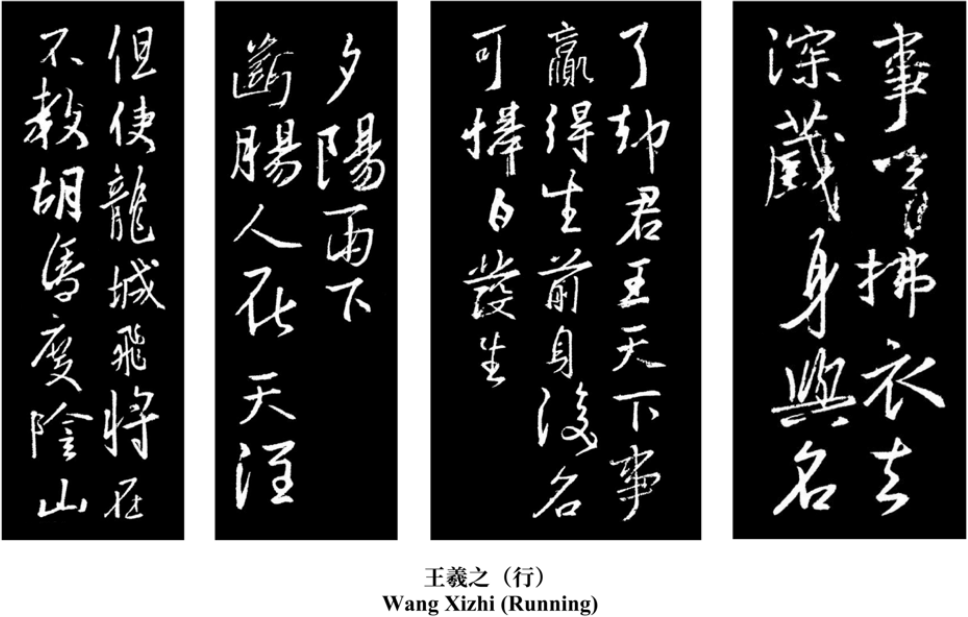}
    \caption{Additional generated samples (page~1 of~6). This collection shows various page-level calligraphy works generated in the style of Wang Xizhi.}
    \label{fig:samples_p1}
\end{figure}

\begin{figure}[H]
    \centering
    \includegraphics[width=0.9\linewidth]{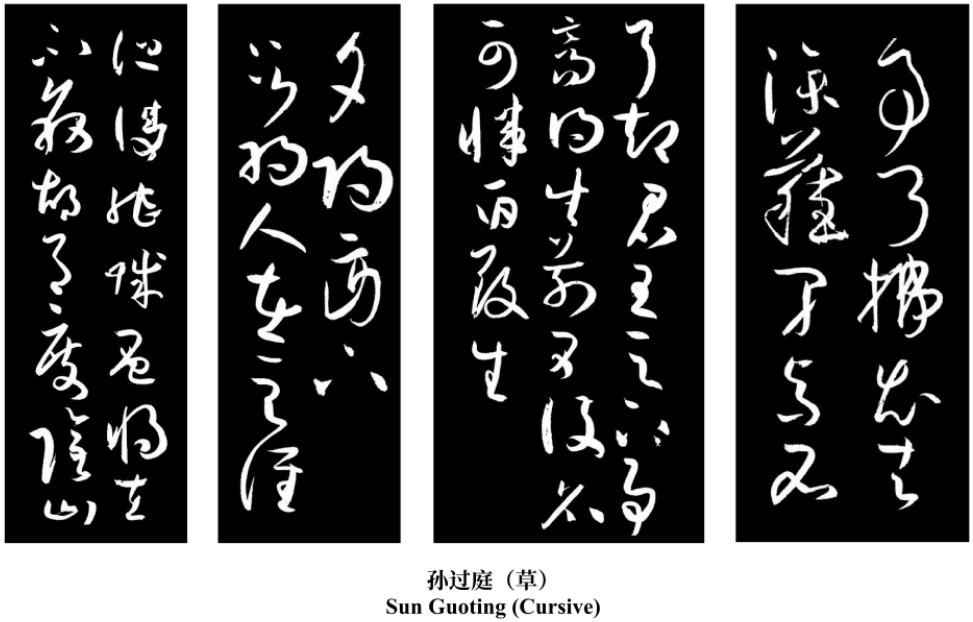}
    \caption{Additional generated samples (page~2 of~6). This collection shows various page-level calligraphy works generated in the style of Sun Guoting. }
    \label{fig:samples_p2}
\end{figure}

\begin{figure}[H]
    \centering
    \includegraphics[width=\linewidth]{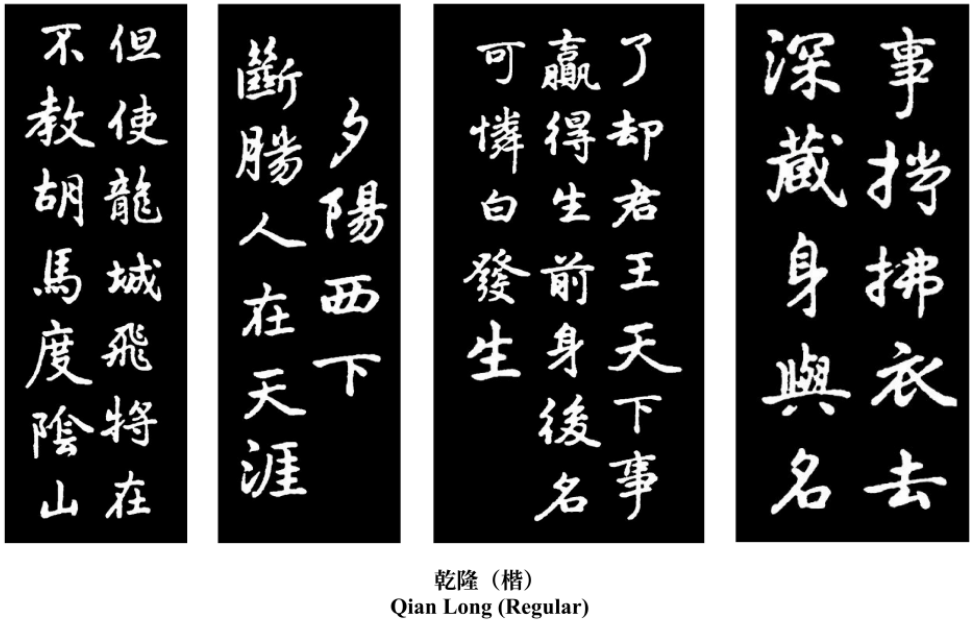}
    \caption{Additional generated samples (page~3 of~6). This collection shows various page-level calligraphy works generated in the style of Qian Long.  }
    \label{fig:samples_p3}
\end{figure}

\begin{figure}[H]
    \centering
    \includegraphics[width=\linewidth]{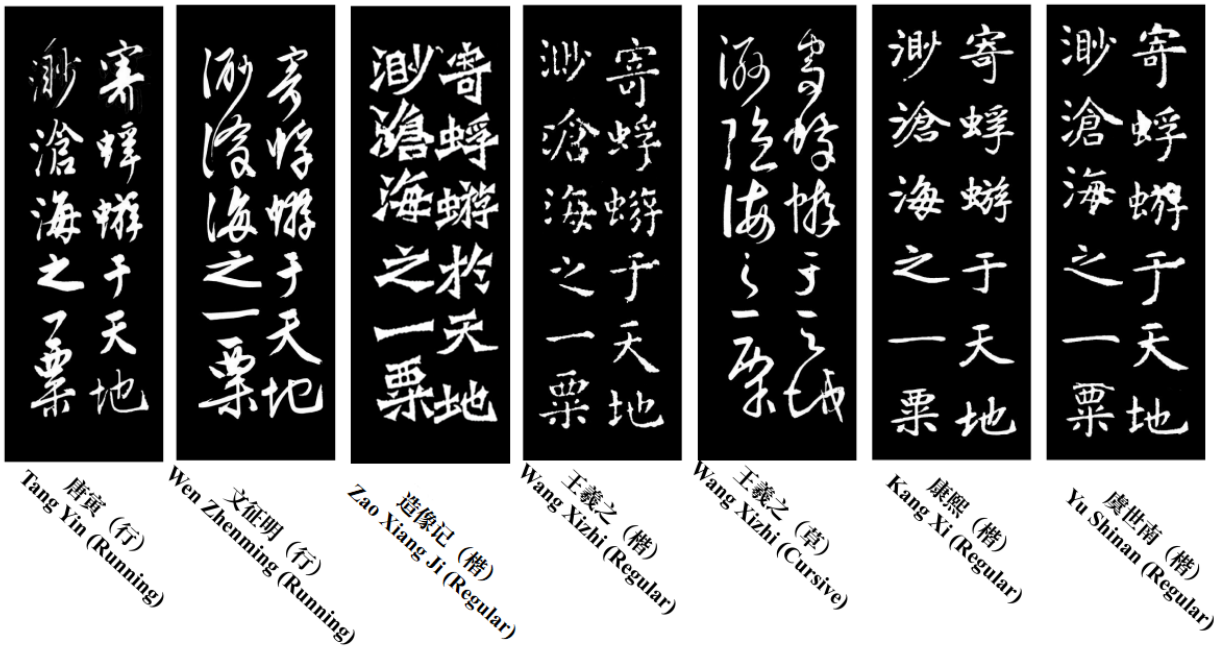}
    \caption{Additional generated samples (page~4 of~6). This collection shows a comparison of different calligraphers and script styles for the same poem. }
    \label{fig:samples_p4}
\end{figure}

\begin{figure}[H]
    \centering
    \includegraphics[width=\linewidth]{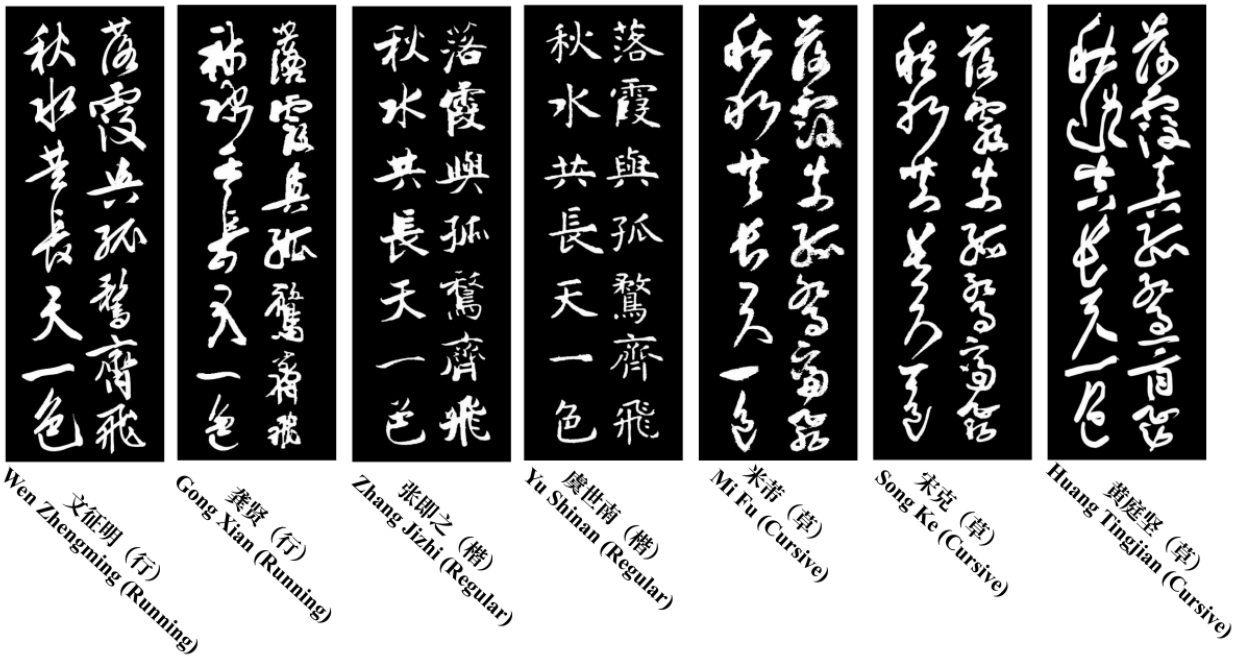}
    \caption{Additional generated samples (page~5 of~6). This collection shows a comparison of various calligraphers and script styles for the same poem. }
    \label{fig:samples_p5}
\end{figure}

\begin{figure}[H]
    \centering
    \includegraphics[width=\linewidth]{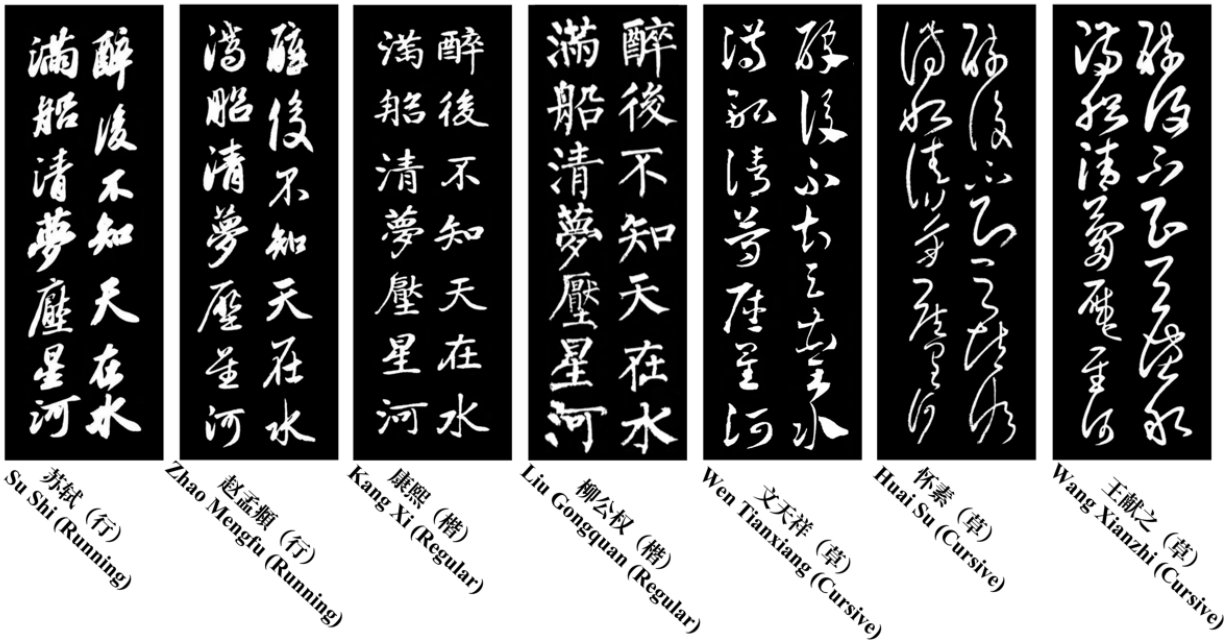}
    \caption{Additional generated samples (page~6 of~6). This final collection shows more examples of different calligraphers and script styles for the same poem. }
    \label{fig:samples_p6}
\end{figure}

Figures~\ref{fig:samples_p1}--\ref{fig:samples_p6} present an extended gallery of CalliMaster outputs spanning a wide range of calligraphers, script types, and textual contents. Each page pairs a different target calligrapher with varied phrases drawn from classical and contemporary Chinese literature, constructing a diverse combination of style and content that stress-tests the model across its full operational envelope.

A key challenge in personalized calligraphy generation is that individual masters differ not only in stroke-level aesthetics---brush pressure, ink loading, and character geometry---but also in global compositional preferences such as inter-character spacing, column width, and page rhythm. CalliMaster addresses both levels simultaneously: it first infers a style-adapted spatial layout from the target calligrapher's visual identity, then generates characters whose stroke dynamics are consistent with that identity. The results confirm that this two-stage design generalizes robustly: as the calligrapher identity changes, the inferred layout shifts accordingly, and the generated brushwork tracks the new stylistic target without degradation in glyph accuracy.

